\documentclass[letterpaper, 10pt, conference]{ieeeconf}      % Use this line for a4 paper

\IEEEoverridecommandlockouts                              % This command is only needed if 
\usepackage{graphics} % for pdf, bitmapped graphics files
\usepackage{epsfig} % for postscript graphics files
\usepackage{subfigure}
\usepackage{amsmath} % assumes amsmath package installed
\usepackage{amssymb}  % assumes amsmath package installed
\usepackage{epstopdf}
\usepackage{lipsum}
\usepackage[colorlinks=true]{hyperref}
\usepackage{lipsum,amsmath,multicol}
\usepackage{float}
\usepackage{algorithm}
\usepackage{algpseudocode}
\usepackage{wrapfig}
\usepackage{sidecap}

\usepackage{amsthm}

\usepackage{comment}
\usepackage{array}
\usepackage{glossaries}
\usepackage[font=small]{caption}
\usepackage{makecell}
\makeglossaries
\newcolumntype{L}[1]{>{\raggedright\let\newline\\\arraybackslash\hspace{0pt}}m{#1}}
\newcolumntype{C}[1]{>{\centering\let\newline\\\arraybackslash\hspace{0pt}}m{#1}}
\newcolumntype{R}[1]{>{\raggedleft\let\newline\\\arraybackslash\hspace{0pt}}m{#1}}
\usepackage{sidecap}
\usepackage{url}
\usepackage[T1]{fontenc}
\usepackage{xcolor}
\usepackage{adjustbox}
\usepackage{tabularx}
\usepackage{multicol, blindtext}
\usepackage{amsmath}

\allowdisplaybreaks

\makeatletter
\def\endthebibliography{%
  \def\@noitemerr{\@latex@warning{Empty `thebibliography' environment}}%
  \endlist
}
\makeatother

\begin{document}
\setlength{\abovedisplayskip}{2.5pt}
\setlength{\belowdisplayskip}{2.5pt}
\setlength{\abovedisplayshortskip}{2.5pt}
\setlength{\belowdisplayshortskip}{2.5pt}
 
\title{Fast Adaptation of Manipulator Trajectories to Task Perturbation By Differentiating through the Optimal Solution}
\author{Shashank Srikanth$^{1}$, Mithun Babu$^{1}$, Houman Masnavi$^{2}$, Arun Kumar Singh$^{2}$, \\ Karl Kruusamäe$^{2}$, K. Madhava Krishna$^{1}$

% \thanks{Houman, Arun and Karl are with the Institute of Technology, University of Tartu. The rest are with IIIT-Hyderabad, India. The work was supported in part by the European Social Fund through IT Academy program in Estonia, smart specialization project with BOLT, CHIST-ERA project InDex and Estonian Centre of Excellence in IT (EXCITE) funded by the European Regional Development Fund.}
% }

\thanks{$^1$Robotics Research Center, KCIS, IIIT Hyderabad, India.}
\thanks{$^2$Institute of Technology, University of Tartu.}
\thanks{$^3$The work was supported in part by the European Social Fund through the IT Academy program in Estonia, smart specialization project with BOLT, CHIST-ERA project InDex, and Estonian Centre of Excellence in IT (EXCITE) funded by the European Regional Development Fund.}
}
\maketitle

%Collision avoidance constraints written for many obstacle shapes (e.g circles, ellipses, etc.) have the so called concave form. As a result, trajectory optimization with such constraints can be efficiently solved to a locally optimal solution through a technique called the convex-concave procedure (CCP), which essentially constructs a series of convex approximations to the original non-convex problem. A larger number of existing works on motion planning leverages this result. 

%For example, some parts of our proposed optimizer can be solved in parallel and has a computational cost equivalent to solving a set of linear equations. The remaining parts of our optimizer have a symbolic solution, i.e, the solution is available as a formula which can be evaluated without any extensive numerical computation.

\begin{abstract}
Joint space trajectory optimization under end-effector task constraints leads to a challenging non-convex problem. Thus, a real-time adaptation of prior computed trajectories to perturbation in task constraints often becomes intractable. Existing works use the so-called warm-starting of trajectory optimization to improve computational performance. We present a fundamentally different approach that relies on deriving analytical gradients of the optimal solution with respect to the task constraint parameters. This gradient map characterizes the direction in which the prior computed joint trajectories need to be deformed to comply with the new task constraints. Subsequently, we develop an iterative line-search algorithm for computing the scale of deformation. Our algorithm provides near real-time adaptation of joint trajectories for a diverse class of task perturbations such as (i) changes in initial and final joint configurations of end-effector orientation-constrained trajectories and (ii) changes in end-effector goal or way-points under end-effector orientation constraints. We relate each of these examples to real-world applications ranging from learning from demonstration to obstacle avoidance. We also show that our algorithm produces trajectories with quality similar to what one would obtain by solving the trajectory optimization from scratch with warm-start initialization. But most importantly, our algorithm achieves a worst-case speed-up of 160x over the latter approach. 
\end{abstract}

% \begin{figure}[H]
% \centering     %%% not \center
% \subfigure{\label{fig:a2}\includegraphics[width=0.47\columnwidth]{figs_icra/demonstration.jpg}}
% \subfigure{\label{fig:b2}\includegraphics[width=0.51\columnwidth]{figs_icra/final_config/pitch_all.pdf}}
% \caption{Plots for final joint configuration perturbation}
% \label{demo}
% \end{figure}

\section{Introduction}
A change in task-specification is often unavoidable in real-world manipulation problems. For example, consider a scenario where a manipulator is handing over an object to a human. The robot's estimate of the goal position can change as it executes its prior computed trajectories. Consequently, it needs to quickly adapt its joint motions to reach the new goal position. In this paper, we model motion planning as a parametric optimization problem wherein the task specifications are encoded in the parameters. In this context, adaptation to a new task requires re-computing the optimal joint trajectories for the new set of parameters. This is a computationally challenging process as the underlying cost functions in typical manipulation tasks are highly non-linear and non-convex \cite{dimitry_manifold_planning}. Existing works leverage the so-called warm-starting technique where prior computed trajectories are used as initialization for the optimization solvers \cite{memory_of_motion_mansard}. However, our extensive experimentation with off-the-shelf optimization solvers such as Scipy-SLSQP \cite{scipy} show it is not sufficient for real-time adaptation of joint trajectories to task perturbations. 

\subsection{Main Idea}
\noindent The proposed work explores an alternate approach based on differentiating the optimal solution with respect to the problem parameters, hereafter referred to as the Argmin differentiation \cite{argmin_paper}. To understand this further, consider the following constrained optimization problem over variable $\boldsymbol{\xi}$ (e.g joint angles) and parameter vector $\textbf{p}$ (e.g end-effector position).

\small
\begin{align}
    \boldsymbol{\xi}^*(\textbf{p})= \arg\min f(\boldsymbol{\xi}, \textbf{p})\qquad \label{cost_ex}\\
    g_i(\boldsymbol{\xi}, \textbf{p}) \leq 0, \forall i = 1,2, \dots n \\
   h_j(\boldsymbol{\xi}, \textbf{p}) = 0, \forall j = 1,2, \dots m
\end{align}
\normalsize

\noindent The optimal solution $\boldsymbol{\xi}^*$ satisfies the following Karush-Kuhn Tucker (KKT) conditions.

\small
\begin{subequations}
\begin{align}
    \nabla f(\boldsymbol{\xi}^*, \textbf{p}) +\sum_i\lambda_i\nabla g_i(\boldsymbol{\xi}^*, \textbf{p})+\sum_j \mu_j \nabla h_j(\boldsymbol{\xi}^*, \textbf{p}) = 0 \label{grad_condition}  \\
    g_i(\boldsymbol{\xi}^*, \textbf{p}) \leq 0, \forall i \label{primal_feas_ineq} \\
    h_j(\boldsymbol{\xi}^*, \textbf{p}) =0 \label{primal_feas_eq} \\
    \lambda_i\geq 0, \lambda_i g_i(\boldsymbol{\xi}^*, \textbf{p})=0, \forall i \label{complimentary} .
\end{align}
\end{subequations}

\normalsize

\noindent The gradients in (\ref{grad_condition}) is taken with respect to $\boldsymbol{\xi}$. The variables $\lambda_i, \mu_j$ are called the Lagrange multipliers. Now, consider a scenario where the optimal solution $\boldsymbol{\xi}^*$ for the parameter $\textbf{p}$ needs to be adapted for the perturbed set  $\overline{\textbf{p}} = \textbf{p}+\Delta \textbf{p}$. As mentioned earlier, one possible approach is to resolve the optimization with $\boldsymbol{\xi}^*$ as the warm-start initialization. Alternately, for $\Delta \textbf{p}$ with a small magnitude, an analytical perturbation model can be constructed. To be precise, we can compute the first-order differential of the r.h.s. of (\ref{grad_condition})-(\ref{complimentary}) to obtain analytical gradients in the following form \cite{hauser_opt_map}, \cite{hauser_bilevel}, \cite{andreas_sensolve_1}.

\small
\begin{align}
    (\nabla_{\textbf{p}}\boldsymbol{\xi^{*}}, \nabla_{\textbf{p}}\lambda_i, \nabla_{\textbf{p}}\mu_i^{*} ) = \textbf{F}(\boldsymbol{\xi}^{*}, \textbf{p}, \lambda_i, \mu_j)
    \label{kkt_grad}
\end{align}
\normalsize

\noindent  Multiplying the gradients with $\Delta \textbf{p}$ gives us an analytical expression for the new solution and Lagrange multipliers corresponding to the perturbed parameter set. \cite{andreas_sensolve_1}.

% Equation (\ref{kkt_grad}) characterizes the direction in which optimal solution changes for a infinitesimal perturbation in the parameters.

\subsection{Contribution}
\noindent \textbf{Algorithmic Contribution:} A critical bottleneck in using gradient map of the form (\ref{kkt_grad}) to compute perturbed solutions is that the mapping between $\Delta \textbf{p}$ and $\lambda_i$ is highly discontinuous. In other words, even a small $\Delta \textbf{p}$ can lead to large changes in the so-called active-set of the inequality constraints. Thus it becomes necessary to develop additional active-set prediction mechanisms \cite{andreas_sensolve_1}. In this paper, we by-pass this complication by instead focusing on the parametric optimization with only bound constraints on the variable set. Argmin differentiation of such problem has a simpler structure, which we leverage to develop a line-search based algorithm to incrementally adopt joint trajectories to larger changes in the parameter/tasks \footnote{To give some example of "large perturbation", our algorithm can adapt the joint trajectories of Franka-Panda arm to perturbation of up to 30 cm in the goal position. This is almost $30 \%$ of the workspace of the Franka arm. }.

\noindent \textbf{Application Contribution}: For the first time, we apply the Argmin differentiation concept to the problem of joint trajectory optimization for the manipulators under end-effector task constraints. We consider a diverse class of cost functions to handle (i) perturbations in joint configurations or (ii) end-effector way-points in orientation-constrained end-effector trajectories. We present an extensive benchmarking of our algorithm's performance as a function of the perturbation magnitude. We also show that our algorithm outperforms the warm-start trajectory optimization approach in computation time by several orders of magnitude while achieving similar quality as measured by task residuals and smoothness of the resulting trajectory.

\subsection{Related Works}
\noindent The concept of Argmin differentiation has been around for a few decades, although often under the name of sensitivity analysis \cite{sensitivity_nlp}, \cite{sensitivity_ipopt}. However, off-late it has seen a resurgence, especially in the context of end-to-end learning of control policies \cite{brandon_diffmpc}, \cite{boyd_l4dc}. Our proposed work is more closely related to those that use Argmin differentiation for motion planning or feedback control. In this context, a natural application of Argmin differentiation is in bi-level trajectory optimization where the gradients of the optimal solution from the lower level are propagated to optimize the cost function at the higher level. This technique has been applied to both manipulation and navigation problems in existing works \cite{hauser_bilevel}, \cite{bi_level_man}. Alternately, Argmin differentiation can also be used for correction of prior-computed trajectories \cite{andreas_sensolve_1}, \cite{l_1_mpc_sensitivity}. 

To the best of our knowledge, we are not aware of any work that uses Argmin differentiation for adaptation of task constrained manipulator joint trajectories. The closest to our approach is \cite{hauser_opt_map} that uses it to accelerate the inverse kinematics problem. Along similar lines, \cite{andreas_sensolve_1} considers a very specific example of perturbation in end-effector goal position. In contrast to these two cited works, we consider a much diverse class of task constraints. Furthermore, our formulation also has important distinctions with \cite{andreas_sensolve_1} at the algorithmic level. Authors in \cite{andreas_sensolve_1} use the log-barrier function for including inequality constraints as penalties in the cost function. In contrast, we note that in the context of the task-constrained trajectory optimization considered in this paper, the joint angle limits are the most critical. The velocity and acceleration constraints can always be satisfied through time-scaling based pre-processing \cite{pham_timescaling}. Thus, by choosing a way-point parametrization for the joint trajectories, we formulate the underlying optimization with just box constraints on the joint angles. This, in turn, allows us to treat this constraint through simple projection (Line 4 in Algorithm \ref{algo_1}) without disturbing the structure of the cost function and the resulting Jacobian and Hessian matrices obtained by Argmin differentiation. 

\section{Proposed Approach}

\subsection{Symbols and Notations}
\noindent We will use lower case normal font letters to represent scalars, while bold font variants will represent vectors. Matrices are represented by upper case bold fonts. The subscript $t$ will be used to denote the time stamp of variables and vectors. The superscript $T$ will represent the transpose of a matrix. 

\subsection{Argmin Differentiation for Unconstrained Parametric Optimization}

\noindent We consider the optimal joint trajectories to be the solution of the following bound-constrained optimization with parameter $\textbf{p}$. 

\small
\begin{subequations}
\begin{align}
    \boldsymbol{\xi}^{*}(\textbf{p}) = \arg\min_{\boldsymbol{\xi}} f(\boldsymbol{\xi}, \textbf{p}) \label{cost_uncon}\\
    \boldsymbol{\xi}_{lb} \leq \boldsymbol{\xi} \leq \boldsymbol{\xi}_{ub}
\end{align}
\end{subequations}
\normalsize

\noindent We are interested in computing the Jacobian of $\boldsymbol{\xi}^{*}(\textbf{p})$ with respect to $\textbf{p}$. If we ignore the bound-constraints for now, we can follow the approach presented in \cite{argmin_paper} to obtain them in the following form.

\small
\begin{align}
\nabla_{\textbf{p}}\boldsymbol{\xi} = -(\nabla_{\xi}^2  f(\boldsymbol{\xi}, \textbf{p}))^{-1} \begin{bmatrix}
\nabla_{\xi, p_1}f(\boldsymbol{\xi}, \textbf{p}), & \dots & \nabla_{\xi, p_n}f(\boldsymbol{\xi}, \textbf{p}) \label{argmin_jac}
\end{bmatrix}
\end{align}
\normalsize
  
\noindent Using (\ref{argmin_jac}), we can derive a local model for the optimal solution corresponding to a perturbation $\Delta \textbf{p}$ as 

\small
\begin{align}
    \boldsymbol{\xi}^{*}(\overline{\textbf{p}}) = \boldsymbol{\xi}^{*}(\textbf{p})+\nabla_{\textbf{p}}\boldsymbol{\xi}^{*} \overbrace{(\overline{\textbf{p}}-\textbf{p})}^{\Delta \textbf{p}}, \label{pert_model}
\end{align}
\normalsize

% To this end, we ignore the bound constraints follow the approach presented in (\cite{argmin_paper}). Since, $\boldsymbol{\xi}^{*}(\textbf{p})$ is optimal, we must have

% \small
% \begin{align}
%     \nabla_{\xi}f(\boldsymbol{\xi}, \textbf{p}) = 0 \label{grad_condition_uucon}
% \end{align}
% \normalsize

% \noindent Differentiating both sides of (\ref{grad_condition_uucon}) with respect to the $i^{th}$ element of $\textbf{p}$, we get:

% \small
% \begin{subequations}
% \begin{align}
%     \frac{d}{dp_i}(\nabla_{\xi}f(\boldsymbol{\xi}, \textbf{p})) = 0 \nonumber \\ 
% \Rightarrow \nabla_{\xi}^2  f(\boldsymbol{\xi}, \textbf{p}))\nabla_{p_i}\boldsymbol{\xi}+\nabla_{\xi, p_i}f(\boldsymbol{\xi}, \textbf{p})) = 0 \nonumber \\
% \nabla_{p_i}\boldsymbol{\xi} = -(\nabla_{\xi}^2  f(\boldsymbol{\xi}, \textbf{p}))^{-1}\nabla_{\xi, p_i}f(\boldsymbol{\xi}, \textbf{p}) \label{arg_min_scalar}
% \end{align}
% \end{subequations}
% \normalsize

% \noindent We can repeat the derivation in (\ref{arg_min_scalar}) with each element of $\textbf{p}$ to get the following Jacobian matrix.

\noindent Intuitively, (\ref{pert_model}) signifies a step of length $\Delta \textbf{p}$ along the gradient direction. However, for (\ref{pert_model}) to be valid, the step-length needs to be small. In other words, the perturbed parameter $\overline{\textbf{p}}$ needs to be in the vicinity of $\textbf{p}$. Although it is difficult to mathematically characterize the notion of "small", in the following, we attempt a practical definition based on the notion of optimal cost.

\newtheorem{definition}{Definition}

\begin{definition}\label{def_pert_model}
A valid $\vert \Delta \textbf{p} \vert$ is one that satisfies the following relationship

\small
\begin{align}
   f(\boldsymbol{\xi}^{*}(\overline{\textbf{p}}= \textbf{p}+\Delta \textbf{p}), \textbf{p}+\Delta \textbf{p})\leq f(\boldsymbol{\xi}^{*}, \textbf{p}+\Delta \textbf{p})
   \label{valid_pert}
\end{align}
\end{definition}
\normalsize

\noindent The underlying intuition in (\ref{valid_pert}) is that the perturbed solution should lead to a lower cost for the parameter $\textbf{p}+\Delta \textbf{p}$
as compared to $\boldsymbol{\xi}^{*}$ for the same perturbed parameter. 

\subsection{Line Search and Incremental Adaption}
\noindent Algorithm \ref{algo_1} couples the concept from the definition (\ref{pert_model}) with a basic line-search to incrementally adapt (\ref{pert_model}) to a large $\Delta \textbf{p}$. The Algorithm begins by initializing the optimal solution  ${^k}\boldsymbol{\xi}$ and the parameter ${^k}\textbf{p}$ with prior values for iteration $k=0$. These variables are then used to initialize the Hessian and Jacobian matrices. The core computations takes place in line 2, wherein we compute the least amount of scaling that needs to be done to step length ${^{k}}\Delta \textbf{p} = {^k}\overline{\textbf{p}}-\textbf{p}$ to guarantee a reduction in the cost. At line 3, we update the optimal solution based on step-length $\eta{^{k}}\Delta \textbf{p}$ obtained in line 2, followed by a simple projection at line 4 to satisfy the minimum and maximum bounds. At line 5, we perform the called forward roll-out of the solution to update the parameter set. For example, if the parameter $\textbf{p}$ models position of the end-effector at final time instant of a trajectory, then line 5 computes how close the ${^{k+1}}\boldsymbol{\xi}^{*}$ takes the end-effector to the perturbed goal position $\overline{\textbf{p}}$. On lines $7$ and $8$, we update the Hessian and the Jacobian matrices based on the updated parameter set and optimal solution.

% the goal position corresponding to the current optimal solution ${^{k+1}}\boldsymbol{\xi}^{*}$. This in turn is used in line 6 to update the ${^{k+1}}\Delta \textbf{p}$.  

% \noindent Equation (\ref{pert_model}) is valid only for a small $\Delta \textbf{p}$.  
% Moreover, it is difficult to characterize this notion of ""

% Although it is difficult to mathematically characterize the magnitude of $\Delta \textbf{p}$ for which (\ref{pert_model}) holds, in the following, we attempt a practical definition based on the notion of optimal cost.

% \newtheorem{definition}{Definition}

% \begin{definition}\label{def_nominal_sol}
% A valid $\Delta \textbf{p}$ is one that satisfies the following relationship

% \begin{align}
%   f(\boldsymbol{\xi}^{*}+\Delta \textbf{p}, \textbf{p}+\Delta \textbf{p})\leq f(\boldsymbol{\xi}^{*}, \textbf{p}+\Delta \textbf{p})
%   \label{valid_pert}
% \end{align}
% \end{definition}

% \noindent The underlying intuition in (\ref{valid_pert}) is that the perturbed solution should lead to a lower cost for the parameter $\textbf{p}+\Delta \textbf{p}$
% as compared to $\boldsymbol{\xi}^{*}$ for the same perturbed parameter. Algorithm \ref{algo_1} couples the above definition with a simple line-search over valid $\textbf{p}$.

\begin{algorithm}[!h]
 \caption{Line-Search Based Joint Trajectory Adaptation to Task Perturbation }\label{algo_1}
    \begin{algorithmic}[1]   
\State Initialize  ${^k}\boldsymbol{\xi}^{*}$ as the solution for the prior parameter ${^k}\textbf{p}$, the Hessian ${^k}\nabla_{\xi}^2  f({^k}\boldsymbol{\xi}, \textbf{p})$, the gradient $\nabla_{\xi, p_i}f({^k}\boldsymbol{\xi}, \textbf{p})$, and ${^k}\Delta \textbf{p} = \overline{\textbf{p}}-{^k}\textbf{p}$

\small
\While {$\eta> 0$ }
\begin{subequations}
\small
\begin{align}
\max \eta\\
f({^k}\boldsymbol{\xi}^{*}(\textbf{p}+\eta\Delta \textbf{p}), \textbf{p}+\Delta \textbf{p})\leq f({^k}\boldsymbol{\xi}^{*}, \textbf{p}+\Delta \textbf{p})
\end{align}
\end{subequations}
\State \begin{align}
    {^{k+1}}\boldsymbol{\xi}^{*} = {^{k}}\boldsymbol{\xi}^{*}+ \eta\nabla_{\textbf{p}}\boldsymbol{\xi}^{*}\Delta {^k}\textbf{p}
 \end{align}
\State \begin{align}
{^{k+1}}\boldsymbol{\xi}^{*} = Project(\boldsymbol{\xi}_{lb}, \boldsymbol{\xi}_{ub})
\end{align} 
\State Update ${^{k+1}}\overline{\textbf{p}}$ = $Forward Roll({^{k+1}}\boldsymbol{\xi}^{*})$
\State Update ${^{k+1}}\Delta \textbf{p} = \overline{\textbf{p}}-{^{k+1}}\textbf{p}$.
\State Update Hessian  $\nabla_{\xi}^2  f({^{k+1}}\boldsymbol{\xi}, {^{k+1}}\textbf{p})$.
\State Update Jacobian $\nabla_{\xi, p}f({^{k+1}}\boldsymbol{\xi}, {^{k+1}}\textbf{p})$
\normalsize
\EndWhile
\end{algorithmic}  
\end{algorithm}
\normalsize
\vspace{-11px}

\section{Task Constrained Joint Trajectory Optimization}
This section formulates various examples of the task-constrained trajectory optimization problem and uses the previous section's results for optimal adaptation of joint trajectories under task perturbation. To formulate the underlying costs, we adopt the way-point parametrization and represent the joint angles at time $t$ as $\textbf{q}_t$. Furthermore, we will use $(\textbf{x}_e(\textbf{q}_t), \textbf{o}_e(\textbf{q}_t))$ to describe the end-effector position and orientation in terms of Euler angles respectively.

% pertaining to trajectory smoothness formulated in the following manner.

% \begin{subequations}
% \begin{align}
%     f(\boldsymbol{\xi}) = \sum_i \frac{1}{2} \textbf{q}_i^T\textbf{A}\textbf{q}_i+\textbf{b}^T\textbf{q}_i \label{cost_task}\\
% \end{align}
% \end{subequations}

% \noindent where the matrices $\textbf{A}$ and $\textbf{b}$  model the squared norm of the first, second and third-order finite difference of the joint positions.

\subsection{Orientation Constrained Interpolation Between Joint Configurations} \label{section:A}
\noindent The task here is to compute an interpolation trajectory between a given initial $\textbf{q}_{0}$ and a final joint configuration $\textbf{q}_m$ while maintaining a specified orientation $\textbf{o}_d$ for the end-effector at all times. We model it through the following cost function.

\small
\begin{align}
    \sum_t f_s(\textbf{q}_{t-k:t})+\left \Vert \begin{matrix}
    \textbf{q}_{t_1}-\textbf{q}_0\\
    \textbf{q}_{t_m}-\textbf{q}_m\\
    \end{matrix}\right \Vert_2^2+\sum_t \Vert \textbf{o}_e(\textbf{q}_t)-\textbf{o}_d\Vert_2^2
    \label{orient_interpol_cost}
\end{align}
\normalsize

\noindent The first term in the cost function models smoothness in terms of joint angles from $t-k$ to $t$ \cite{toussaint_GN}. For example, for $k = 1$, the smoothness is defined as the first-order finite difference of the joint positions at subsequent time instants. Similarly, $k= 2, 3$, will model higher order smoothness through second and third-order finite differences respectively. We consider all three finite-differences in our smoothness cost term. The second term ensures that the interpolation trajectory is close to the given initial and final points. The final term in the cost function maintains the required orientation of the end-effector.

We can shape (\ref{orient_interpol_cost}) in the form of (\ref{cost_uncon}) by defining $\boldsymbol{\xi} = (\textbf{q}_{t_1}, \textbf{q}_{t_2}, \dots \textbf{q}_{t_m})$. The bounds will correspond to the maximum and minimum limits on the joint angles at each time instant. We define the parameter set as $\textbf{p} = (\textbf{q}_0, \textbf{q}_m)$. That is, we are interested in computing the adaptation when either or both of $\textbf{q}_{0}$ and $\textbf{q}_m$ gets perturbed.

\begin{figure}[!h]
\centering     %%% not \center
    \subfigure{\label{fig:glass1}\includegraphics[width=0.45\columnwidth]{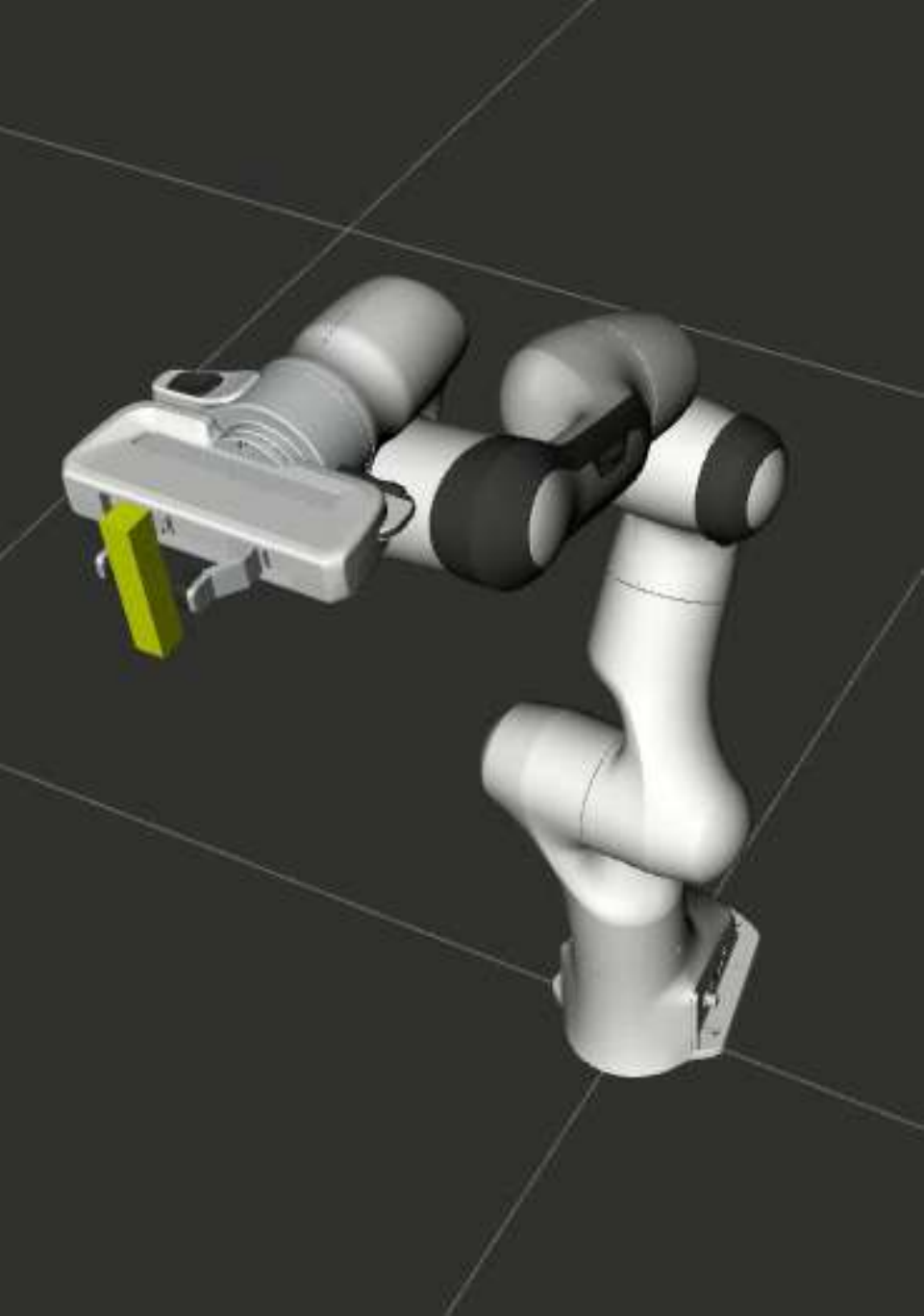}}
    \subfigure{\label{fig:glass2}\includegraphics[width=0.45\columnwidth]{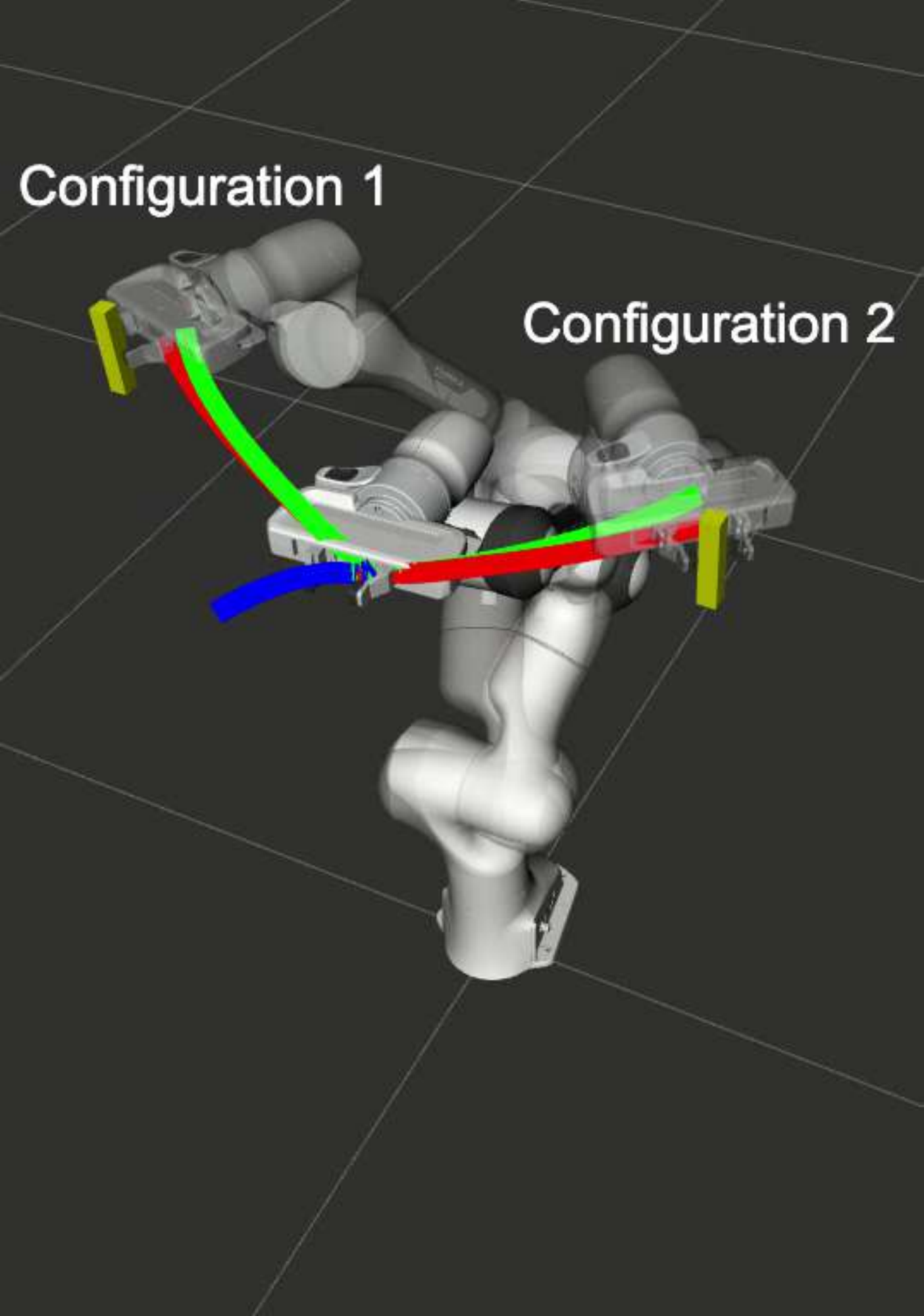}}
    \caption{ Prior trajectory shown in blue is used to adapt the joint motions to move towards two different final joint configurations while maintaining the horizontal orientation of the end-effector at all times. }
    \label{orient_interpol}
\end{figure}
\vspace{-11px}

\begin{figure}[!h]
\centering     %%% not \center
    \subfigure{\label{fig:glass1}\includegraphics[width=0.48\columnwidth]{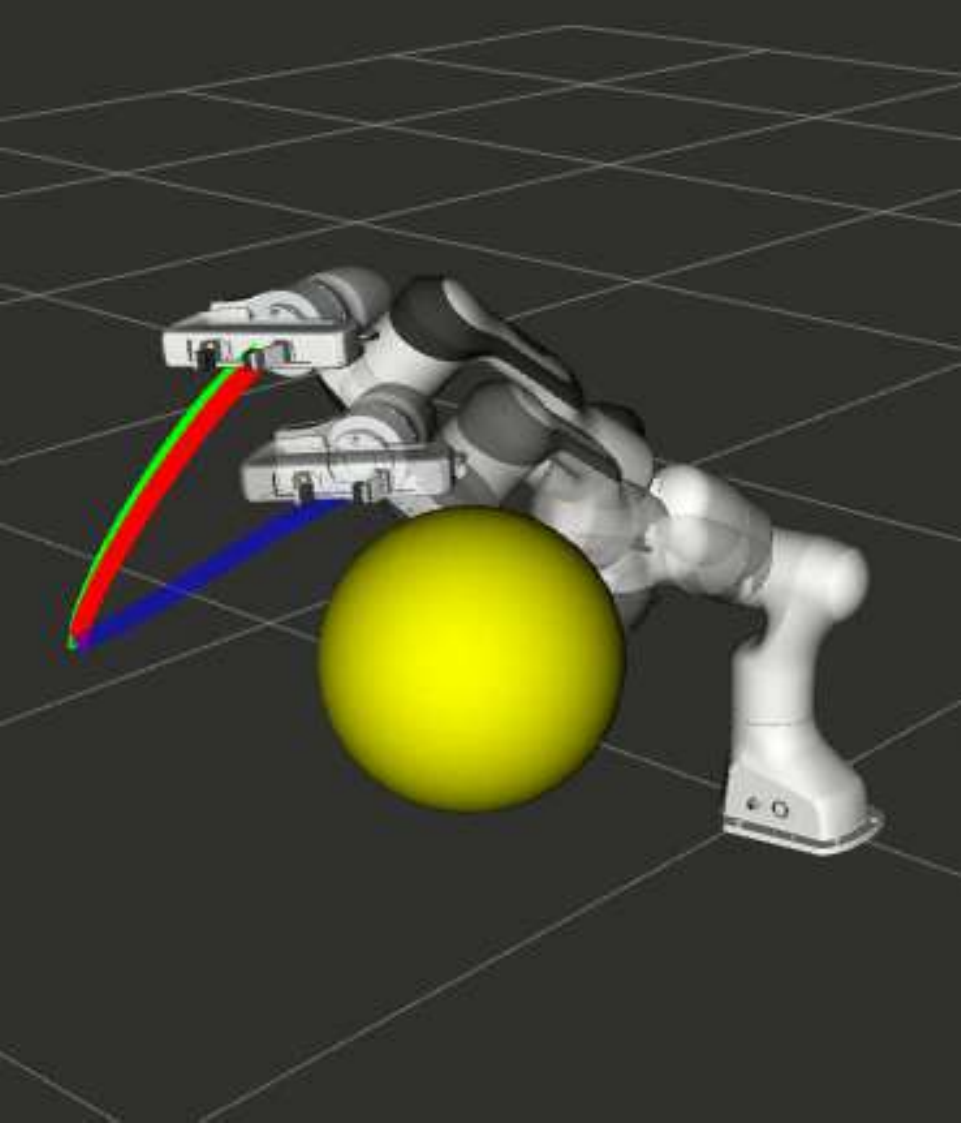}}
    \subfigure{\label{fig:glass2}\includegraphics[width=0.49\columnwidth]{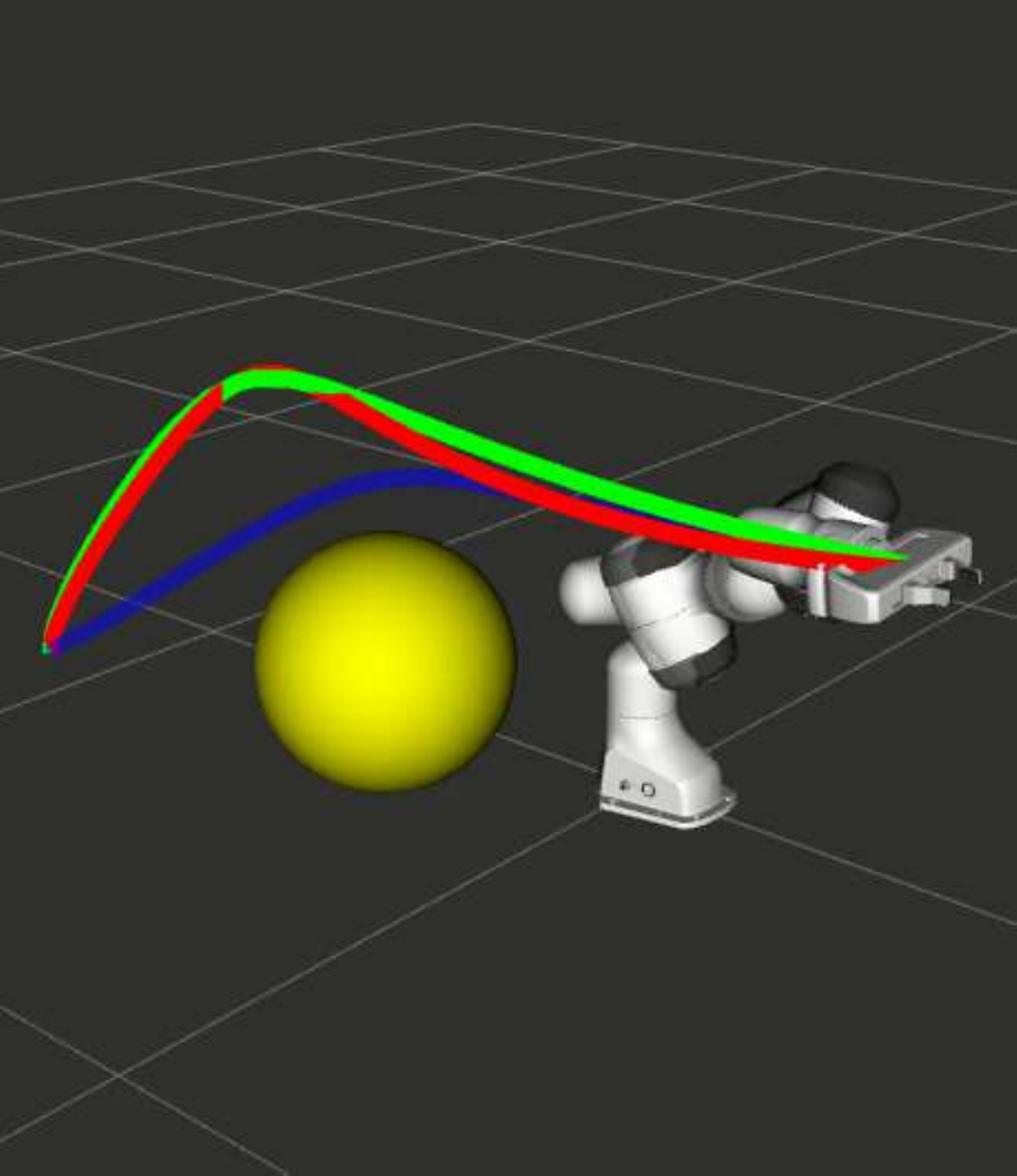}}
    \caption{Collision avoidance by perturbing the mid-point of the prior computed end-effector trajectory.}
    \label{via_point}
\end{figure}
\vspace{2px}

\subsubsection{Applications} Adaptation of $\boldsymbol{\xi}^{*}$ of (\ref{orient_interpol_cost}) for different $\textbf{q}_0, \textbf{q}_m$ has applications in learning from demonstration setting where the human just provides the information about the initial and/or final joint configuration and the manipulator then computes a smooth interpolation trajectory between the boundary configurations by adapting a prior computed trajectory. 

% The demonstration itself can be done by either through kinesthetic teaching or by san interactive device. For the latter, Robot Operating System already provides an off-the-shelf interface where the user can interactively move the manipulator to a desired end-configuration. 

Fig. \ref{orient_interpol} presents an example of adaptation discussed above. The prior computed trajectory is shown in blue. This is then adapted to two different final joint configurations. The trajectory computed through Algorithm \ref{algo_1} is shown in green while that obtained by resolving the optimization problem (with warm-starting) is shown in red. 

\subsection{Orientation-Constrained Trajectories Through Way-Points}\label{section:B}
\noindent The task in this example is to make the end-effector move though given way-points while maintaining the orientation at $\textbf{o}_d$. Let $\textbf{x}_{d_t}$ represent the desired way-point of the end-effector at time $t$. Thus, we can formulate the following cost function for the current task.

\small
\begin{align}
    \sum_t f_s(\textbf{q}_{t-k:t})+\sum_t \Vert \textbf{o}_e(\textbf{q}_t)-\textbf{o}_d\Vert_2^2+\sum_t \Vert \textbf{x}_{e}(\textbf{q}_t)-\textbf{x}_{d_t}\Vert_2^2
    \label{pos_interpol_cost}
\end{align}
\normalsize

\noindent The first two terms in the cost function are same as the previous example. The changes appear in the final term which minimizes the $l_2$ norm of the distance of the end-effector with the desired way-point. The defintion of $\boldsymbol{\xi}$ remains the same as before. However, the parameter set  is now defined as $\textbf{p}=(\textbf{x}_{d_1}, \textbf{x}_{d_2}, \dots \textbf{x}_{d_m} )$.

% \noindent \textbf{Special Case:} If the entire end-effector trajectory needs to be perturbed, defining the parameter set as individual way-points is inefficient. For example, if we define the trajectory through 50 way-points, then the parameter set will be a 150-dimensional vector. This in turn, increases the computation cost of obtaining $\nabla_{p_i, \xi}f(\boldsymbol{\xi}, \textbf{p}) $. Thus, we instead adopt the following approach. We fit a bezier-curve to both the prior and perturbed end-effector trajectory and define our parameter set as the coefficients of the Bezier-curve. In this approach, we can represent the trajectory perturbation in terms of changes in the coefficients.

\subsubsection{Application} \textbf{Collision Avoidance} As shown in Fig. \ref{via_point}, a key application of the adaptation problem discussed above is in collision avoidance. A reactive planner such as \cite{reactive_torsten} can provide new via-points for the manipulator to avoid collision. Our Algorithm \ref{algo_1} can then use the cost function (\ref{pos_interpol_cost}) to adapt prior trajectory shown in blue to that shown in green. For comparison, the trajectory obtained with resolve of the trajectory optimization is shown in red.

\begin{figure}[!h]
\centering     %%% not \center
    \subfigure{\label{fig:glass1}\includegraphics[width=0.48\columnwidth]{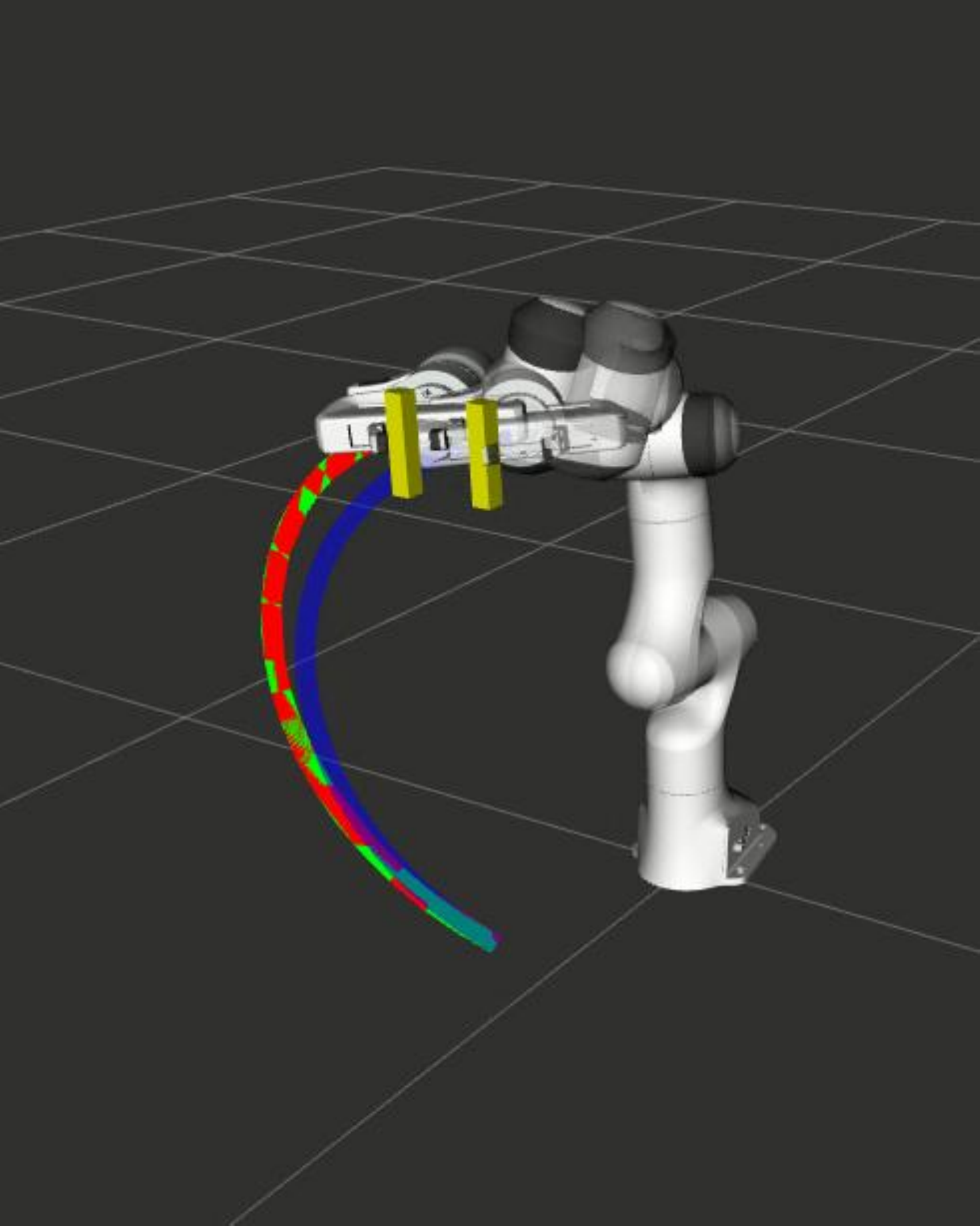}}
    \subfigure{\label{fig:glass2}\includegraphics[width=0.48\columnwidth]{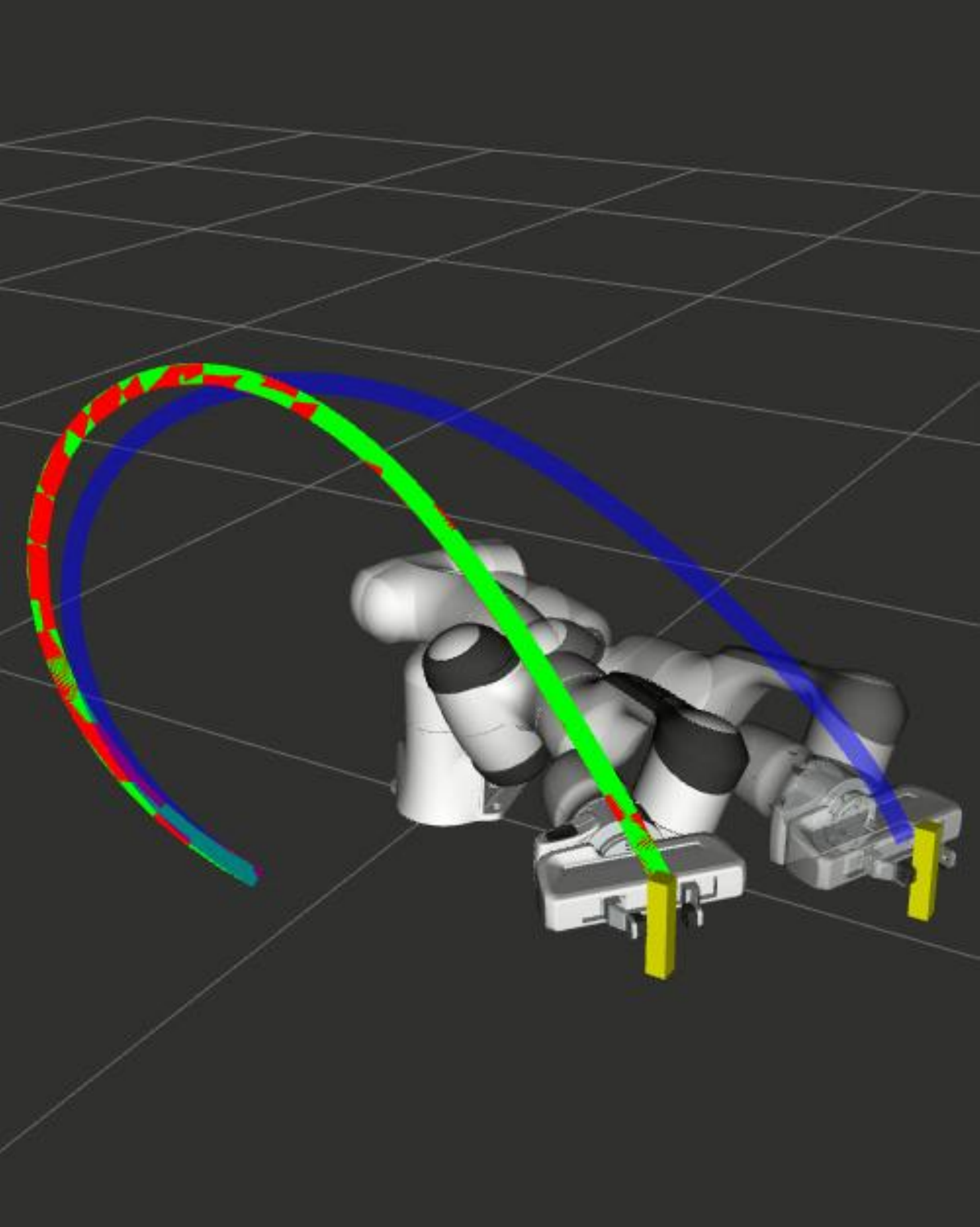}}
    \caption{Perturbation in the final position of the end-effector. }
    \label{final_pos}
\end{figure}

\noindent \textbf{Human-Robot Handover:} Algorithm \ref{algo_1} with cost function (\ref{pos_interpol_cost}) also finds application in human-robot handover tasks. An example is shown in Fig. \ref{final_pos}, where the manipulator adapts the prior trajectory (blue) to a new estimate of the handover position. As before, the trajectory obtained with Algorithm \ref{algo_1} is shown in green, while the one shown in red corresponds to a re-solve of the trajectory optimization with warm-start initialization.

% \begin{figure}[!h]
% \centering     %%% not \center
%     \subfigure{\label{fig:glass1}\includegraphics[width=0.48\columnwidth]{figs_icra/video_plots/S_Traj/S_initial Cropped-converted.pdf}}
%     \subfigure{\label{fig:glass2}\includegraphics[width=0.48\columnwidth]{figs_icra/video_plots/S_Traj/S_final Cropped-converted.pdf}}
%     \caption{Joint Trajectories computed for drawing a letter "S" (blue) has been adapted to draw an enlarged and displaced version of the same letter.}
%     \label{s_curve}
% \end{figure}

% \vspace{-22px}

% \noindent \textbf{Drawing/Painting:} Manipulators are often used in painting/drawing tasks and in this context works like \cite{guassian_dynamical} provides efficient algorithms for reproducing scaled and translated variants of the human demonstration. However, these demonstrations are conventionally at task-level and instantaneous inverse-kinematics is used to compute the necessary joint motions. A better alternative is to solve the task constrained trajectory optimization that leads to smoother joint trajectories, especially in the presence of redundant degrees of freedom \cite{joint_ik_paper}. Our Algorithm \ref{algo_1} has a strong utility in this regard. As shown in Fig. \ref{s_curve}, it can adapt prior computed joint trajectories to trace a scaled and translated variant of the prior end-effector trajectory, the letter "S" in this example.

% \subsection{Motion Planning with Goal-Set Constraints}
% \noindent This task is inspired by \cite{chomp_constrained} that shows the benefit of allowing the manipulator to choose the end-effector goal-position from a set rather than fixing it to one specific value. We define our goal set as a 2D circle around a given position $(x_f, y_f, z_f)$.

% \begin{align}
%     \textbf{x}_g = \begin{bmatrix}
%     x_f+r\cos(\theta)\\
%     y_f+r\cos(\theta)\\
%     z_f
%     \end{bmatrix}
% \end{align}

% \begin{align}
%     \sum_t f_s(\textbf{q}_{t-k:t})+\Vert \textbf{x}_e(\textbf{q}_{t_m})-\textbf{x}_g\Vert_2^2.
% \end{align}
\section{Benchmarking}

\subsection{Implementation Details}
\noindent The objective of this section is to compare the trajectories computed by Algorithm \ref{algo_1} with that obtained by re-solving the trajectory optimization for the perturbed parameters with warm-start initialization. We consider the same three benchmarks presented in Fig. \ref{orient_interpol}-\ref{final_pos}  implemented on a 7dof Franka Panda Arm, but for a diverse range of perturbations magnitude. For each benchmark, we created a data set of 180 trajectory by generating random perturbations in the task parameters. For the benchmark of Fig. \ref{orient_interpol}, the parameters are the joint angles but in the following we use the forward kinematics to derive equivalent representation for the parameters in terms of end-effector position values. 

\begin{figure}[!h]
\centering     %%% not \center
\subfigure{\label{fig:a2}\includegraphics[width=0.47\columnwidth]{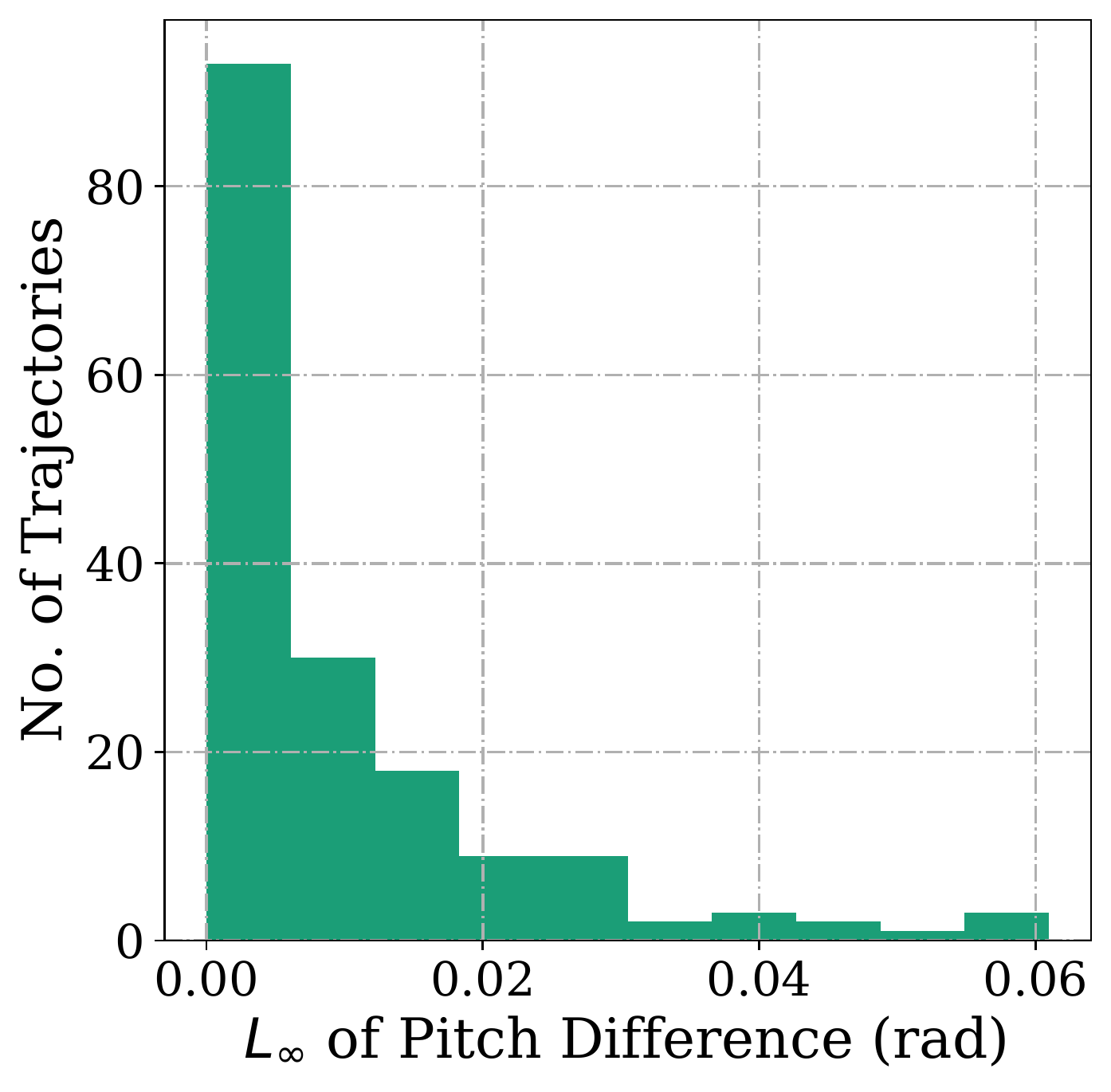}}
\subfigure{\label{fig:b2}\includegraphics[width=0.51\columnwidth]{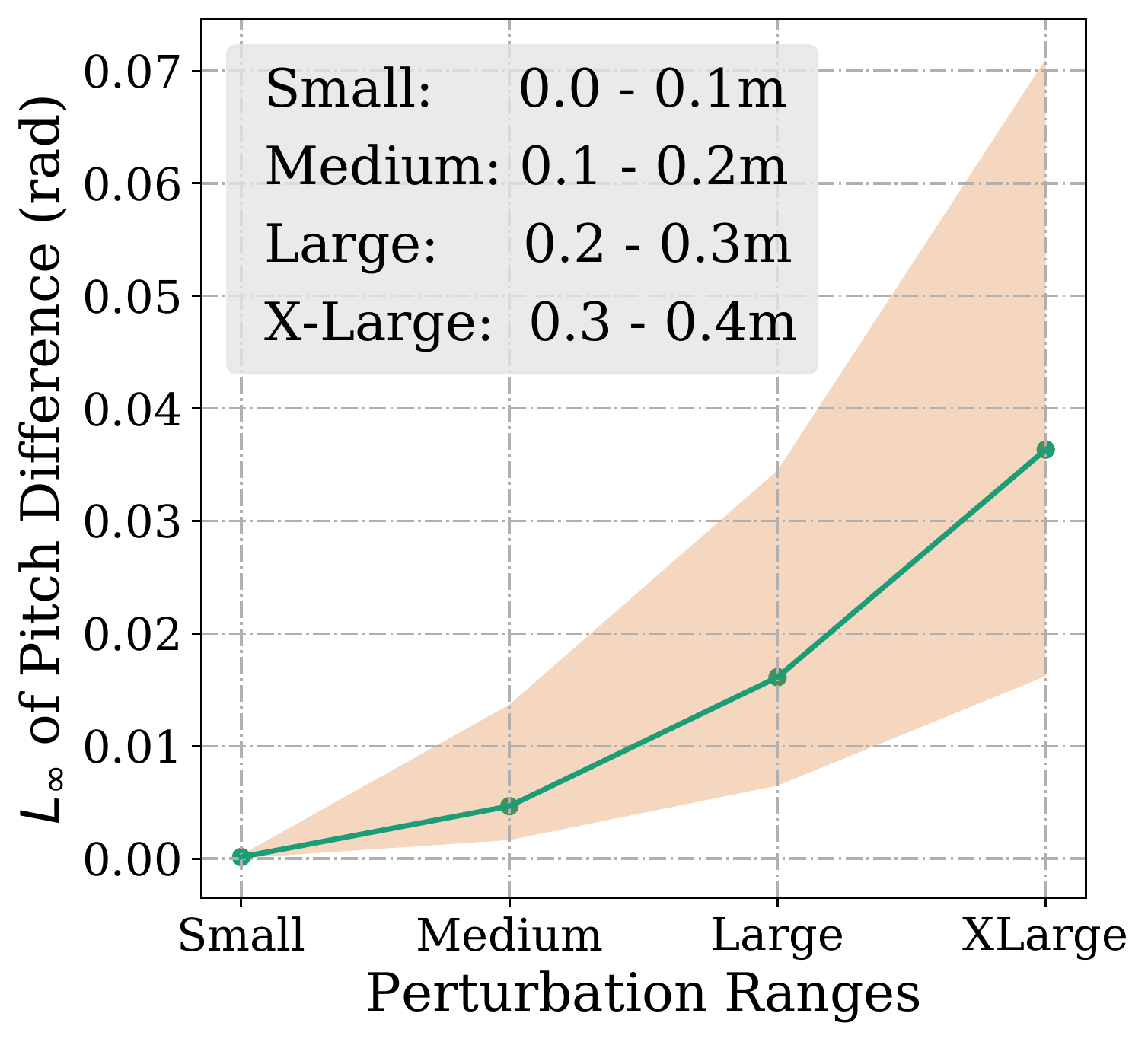}}
\subfigure{\label{fig:c2}\includegraphics[width=0.48\columnwidth]{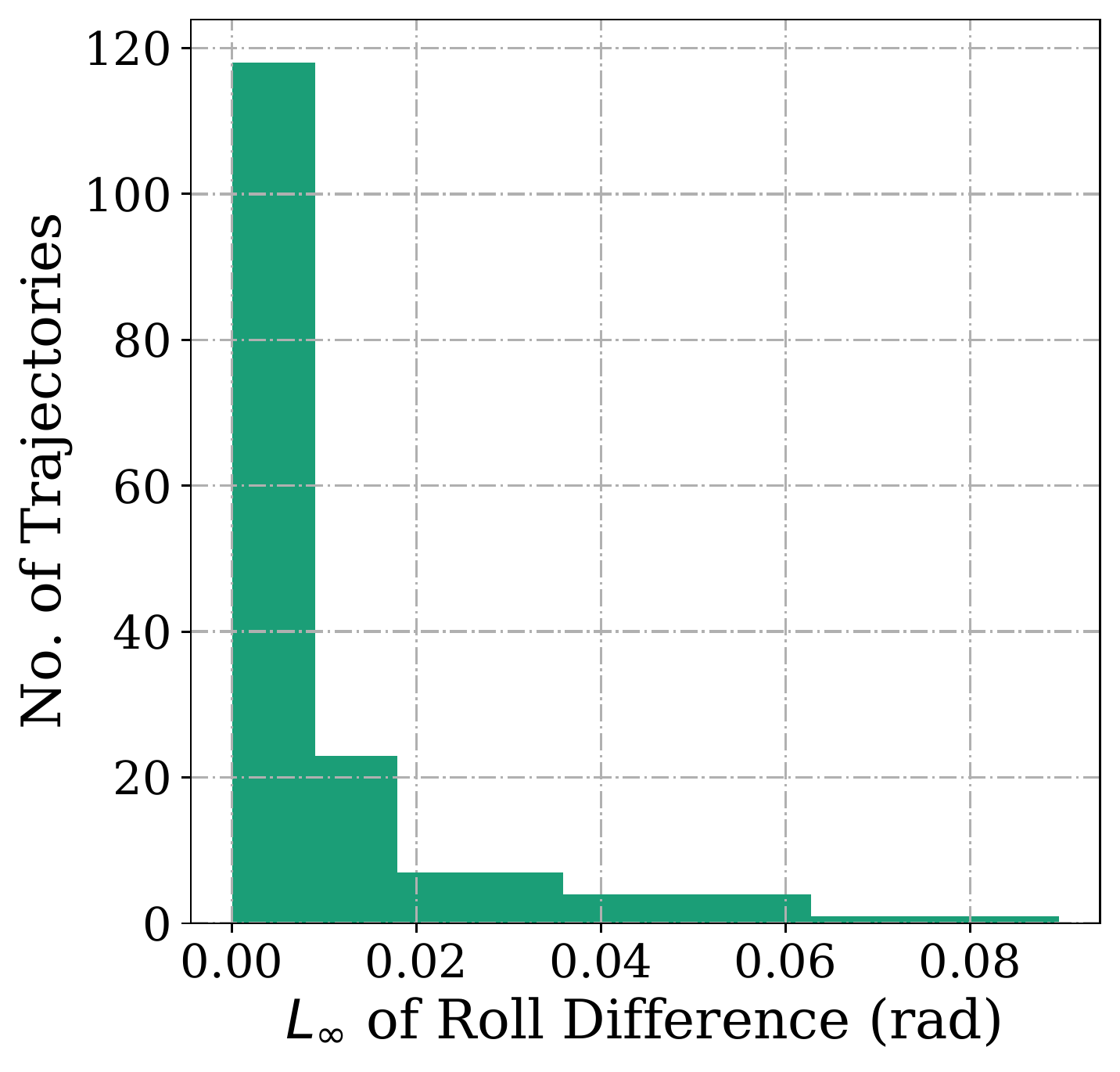}}
\subfigure{\label{fig:d2}\includegraphics[width=0.50\columnwidth]{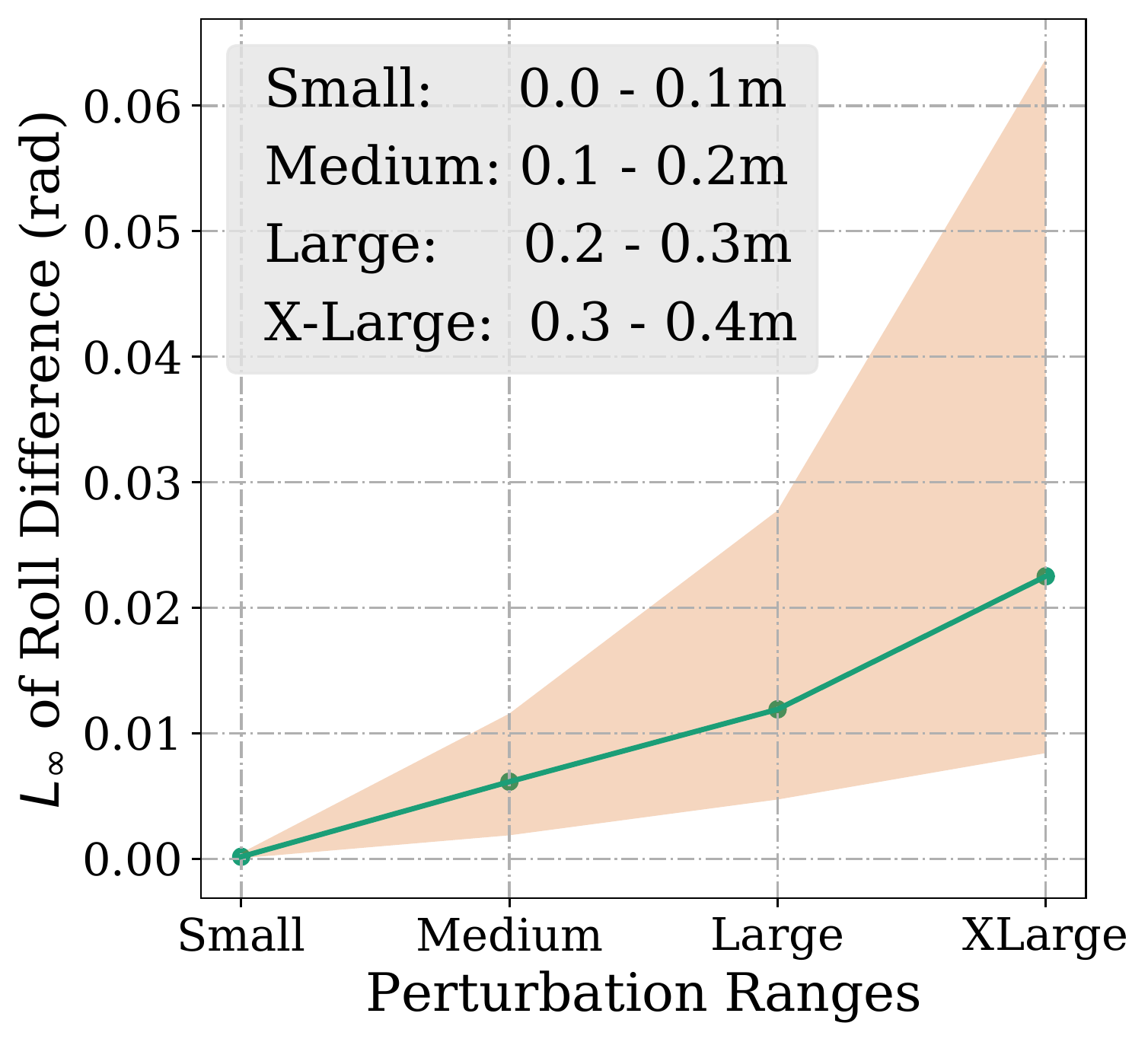}}
\subfigure{\label{fig:c2}\includegraphics[width=0.48\columnwidth]{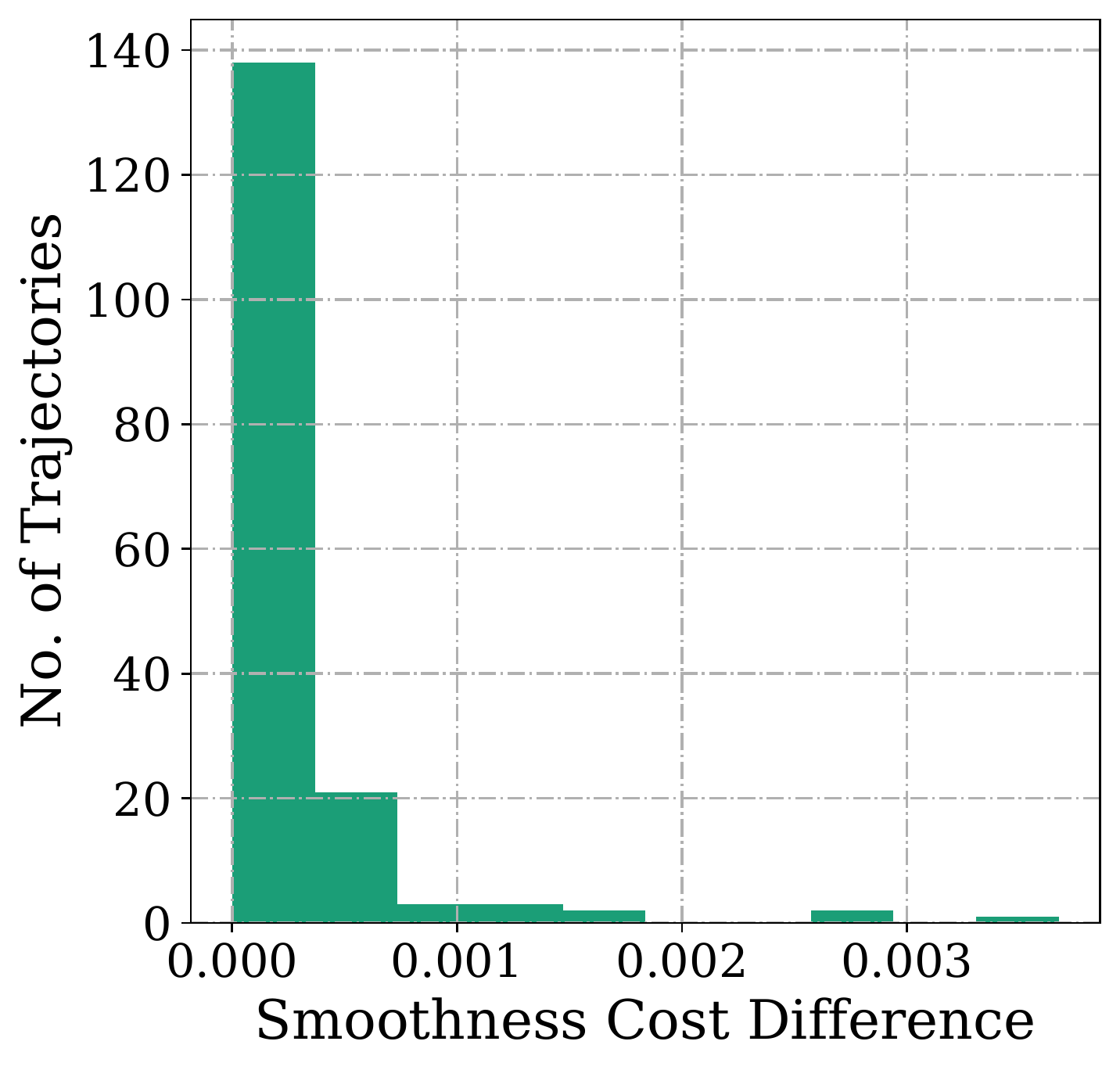}}
\subfigure{\label{fig:d2}\includegraphics[width=0.50\columnwidth]{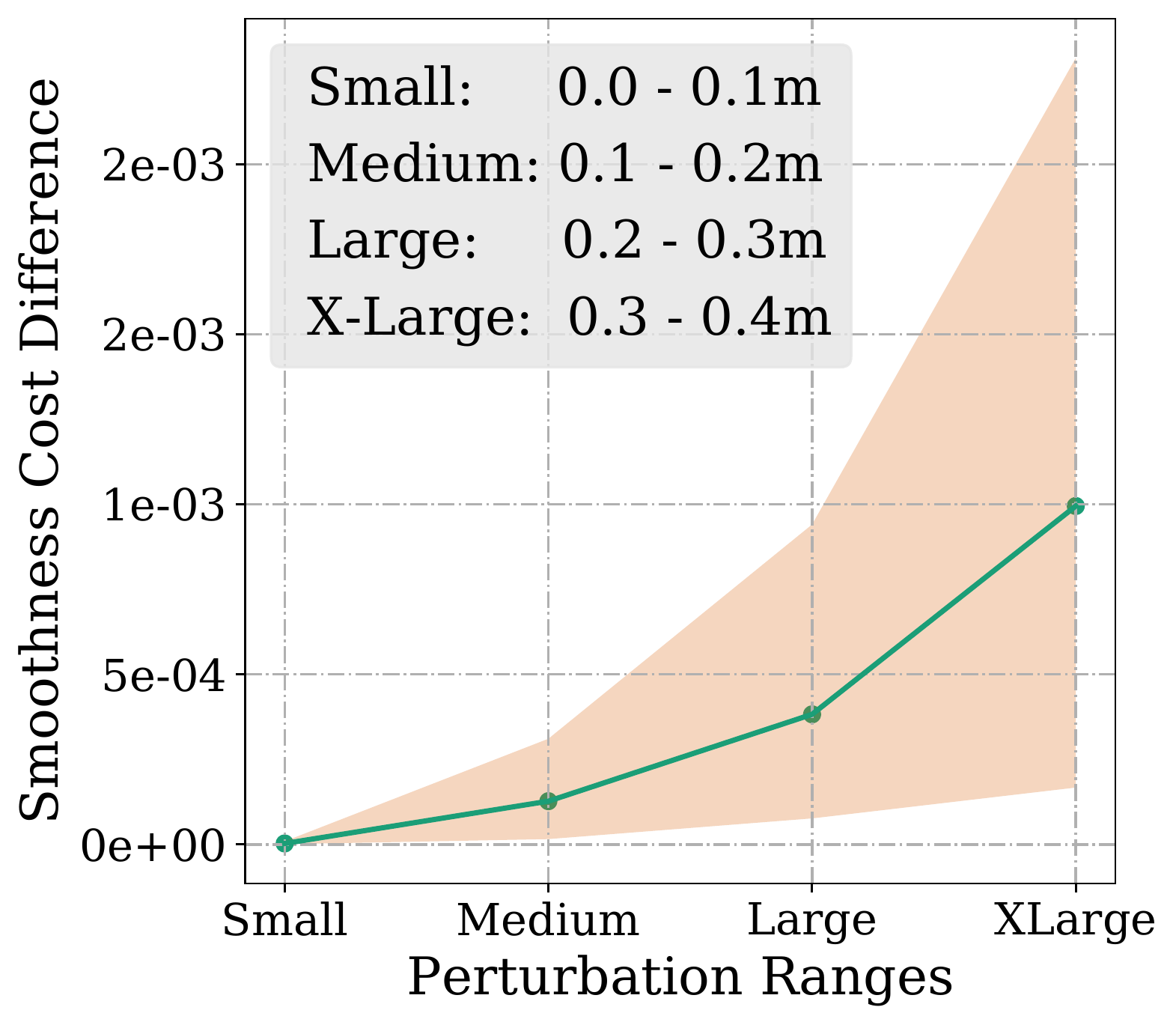}}
\subfigure{\label{fig:e2}\includegraphics[width=0.47\columnwidth]{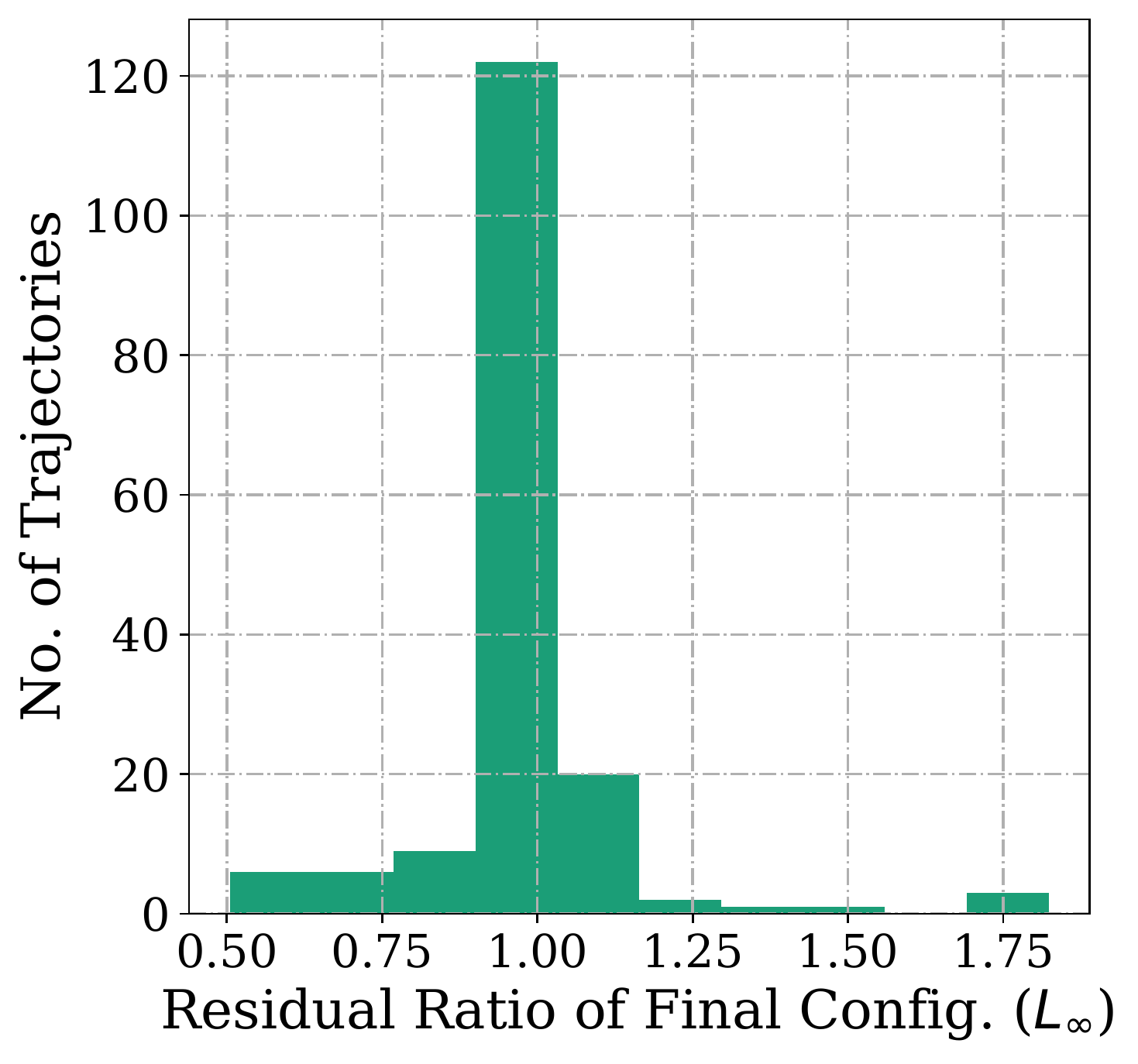}}
\subfigure{\label{fig:f2}\includegraphics[width=0.51\columnwidth]{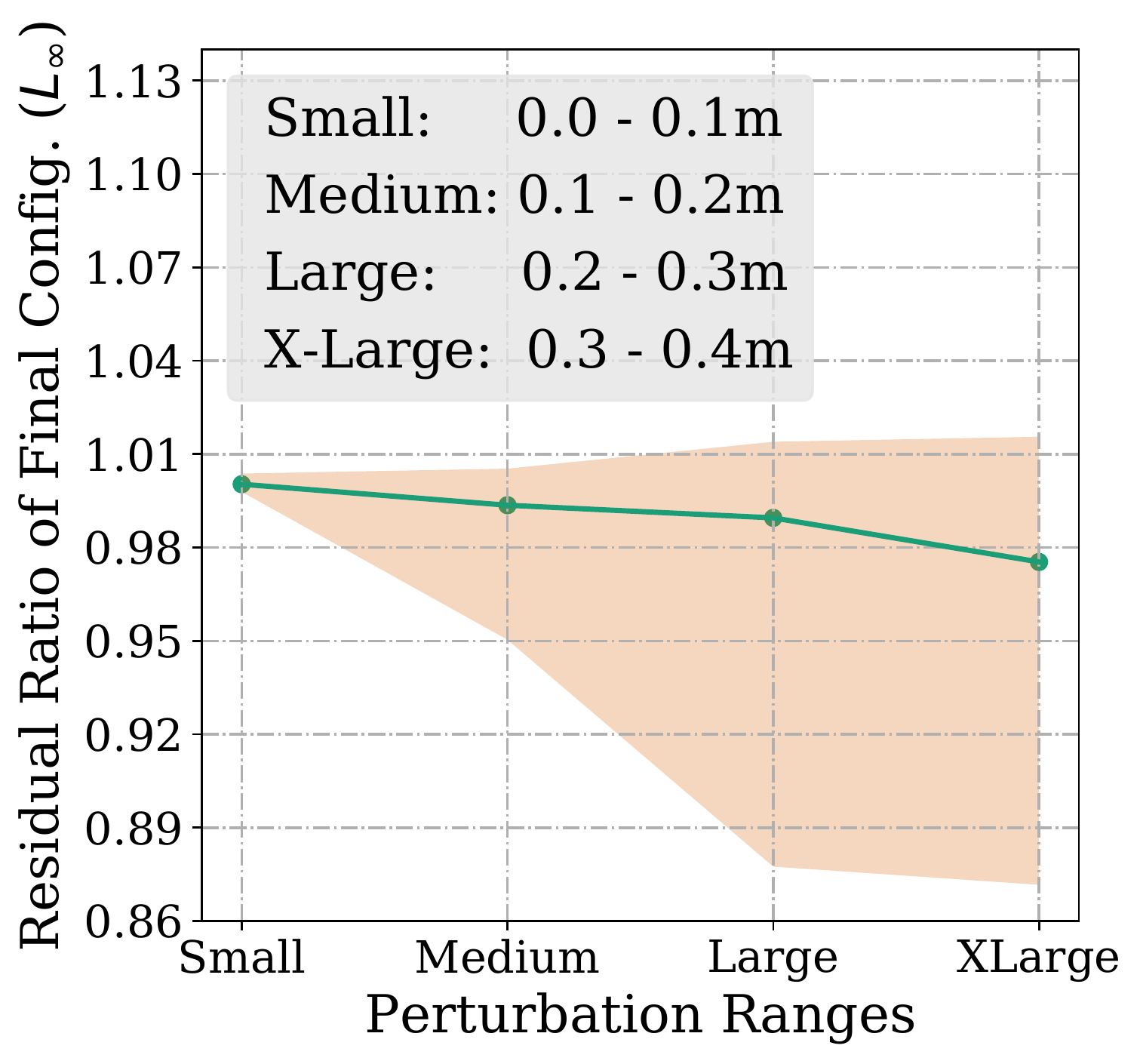}}
\caption{Performance of Algorithm \ref{algo_1} for different perturbation ranges on the benchmark of Fig. \ref{orient_interpol} that involves perturbing the final joint configuration (recall cost function (\ref{orient_interpol_cost})). Note that the perturbation in final joint is converted to position values by forward kinematics. The left column show the histogram of orientation, smoothness and task residual ratio metrics for the medium range perturbation. The right column quantifies the metrics for different perturbation ranges. }
\label{final_joint_pert}
\end{figure}

\begin{figure}[!h]
\centering     %%% not \center
\subfigure{\label{fig:a4}\includegraphics[width=0.47\columnwidth]{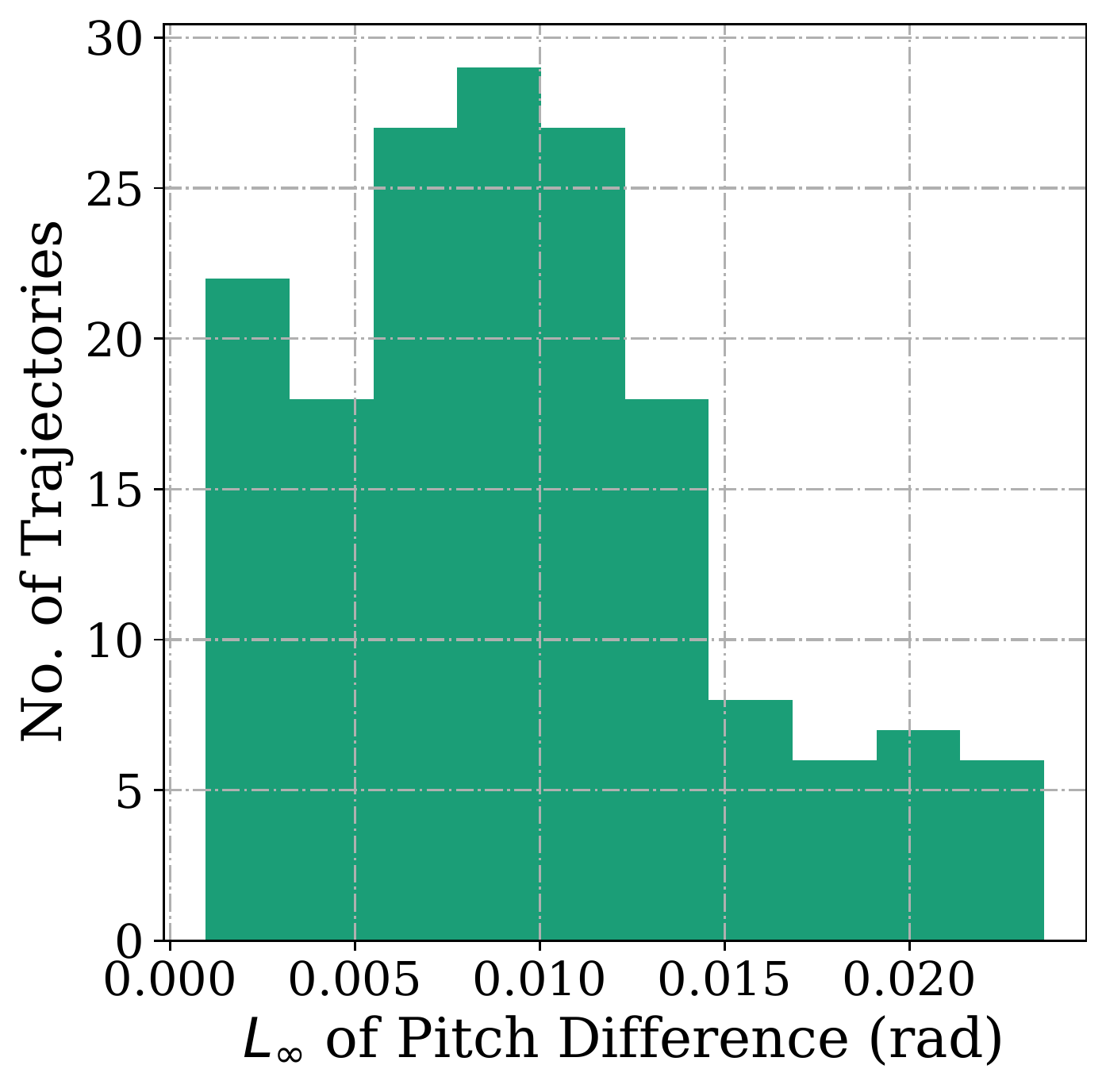}}
\subfigure{\label{fig:b4}\includegraphics[width=0.51\columnwidth]{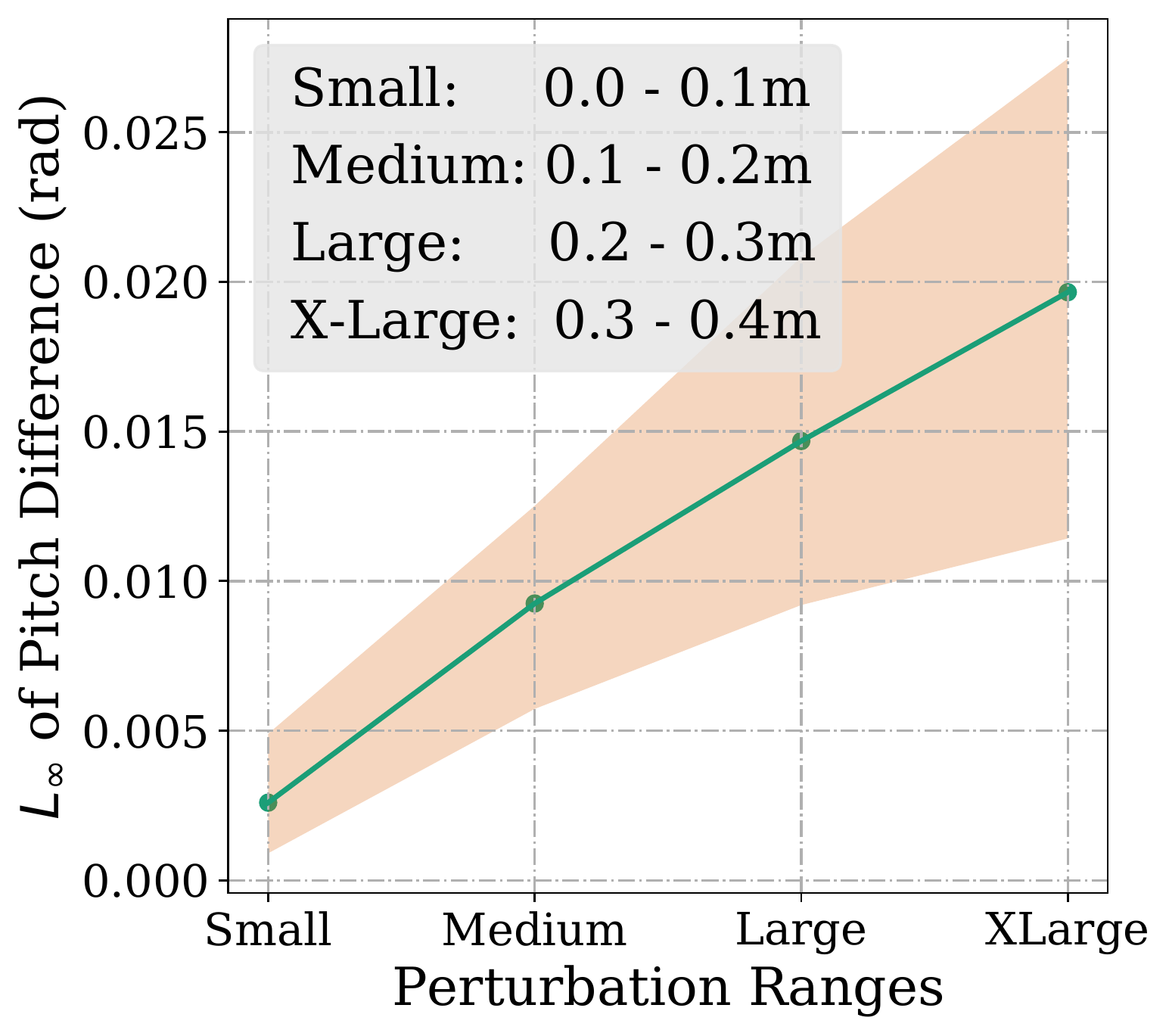}}
\subfigure{\label{fig:c4}\includegraphics[width=0.47\columnwidth]{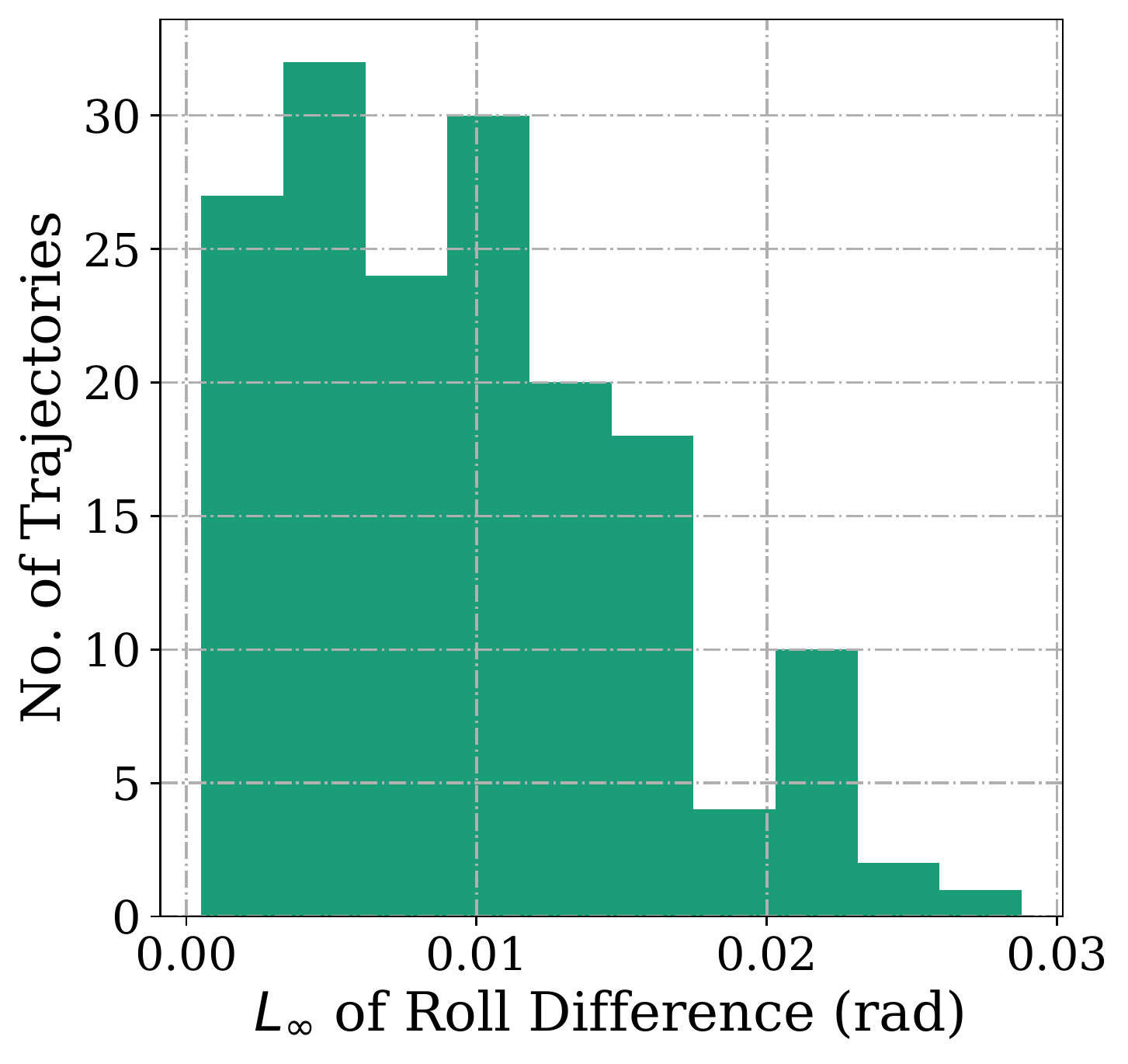}}
\subfigure{\label{fig:d4}\includegraphics[width=0.51\columnwidth]{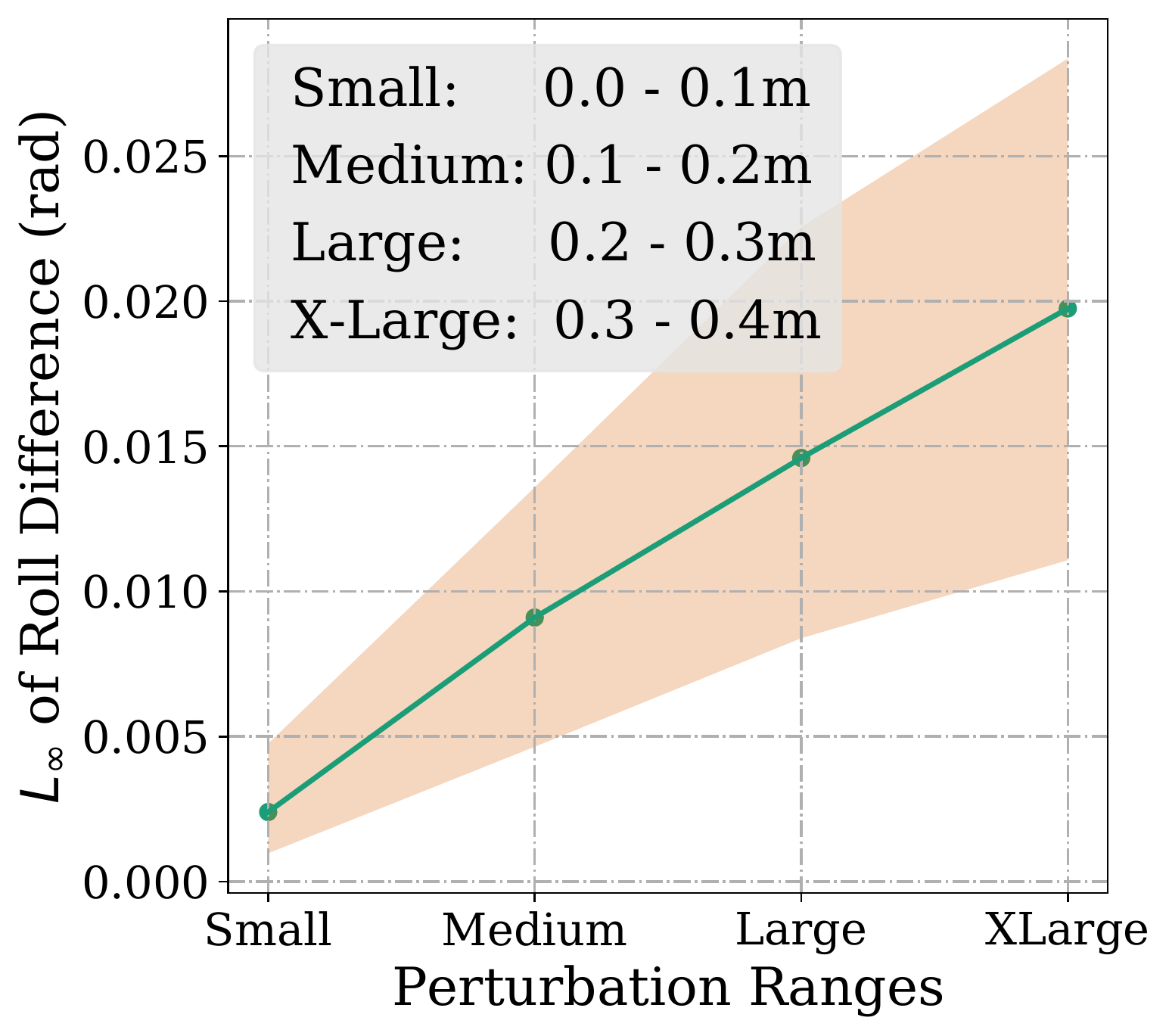}}
\subfigure{\label{fig:c4}\includegraphics[width=0.47\columnwidth]{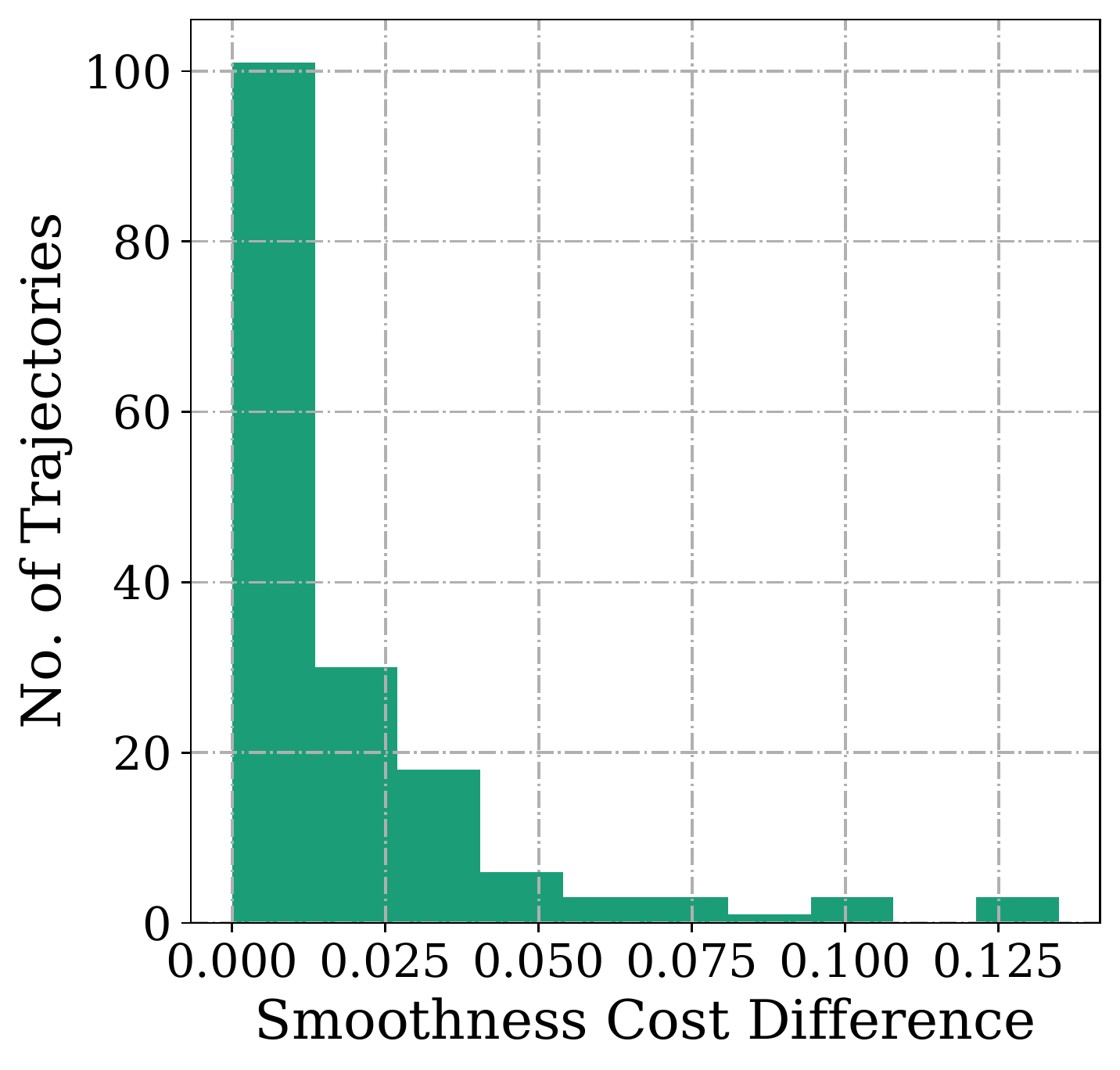}}
\subfigure{\label{fig:d4}\includegraphics[width=0.51\columnwidth]{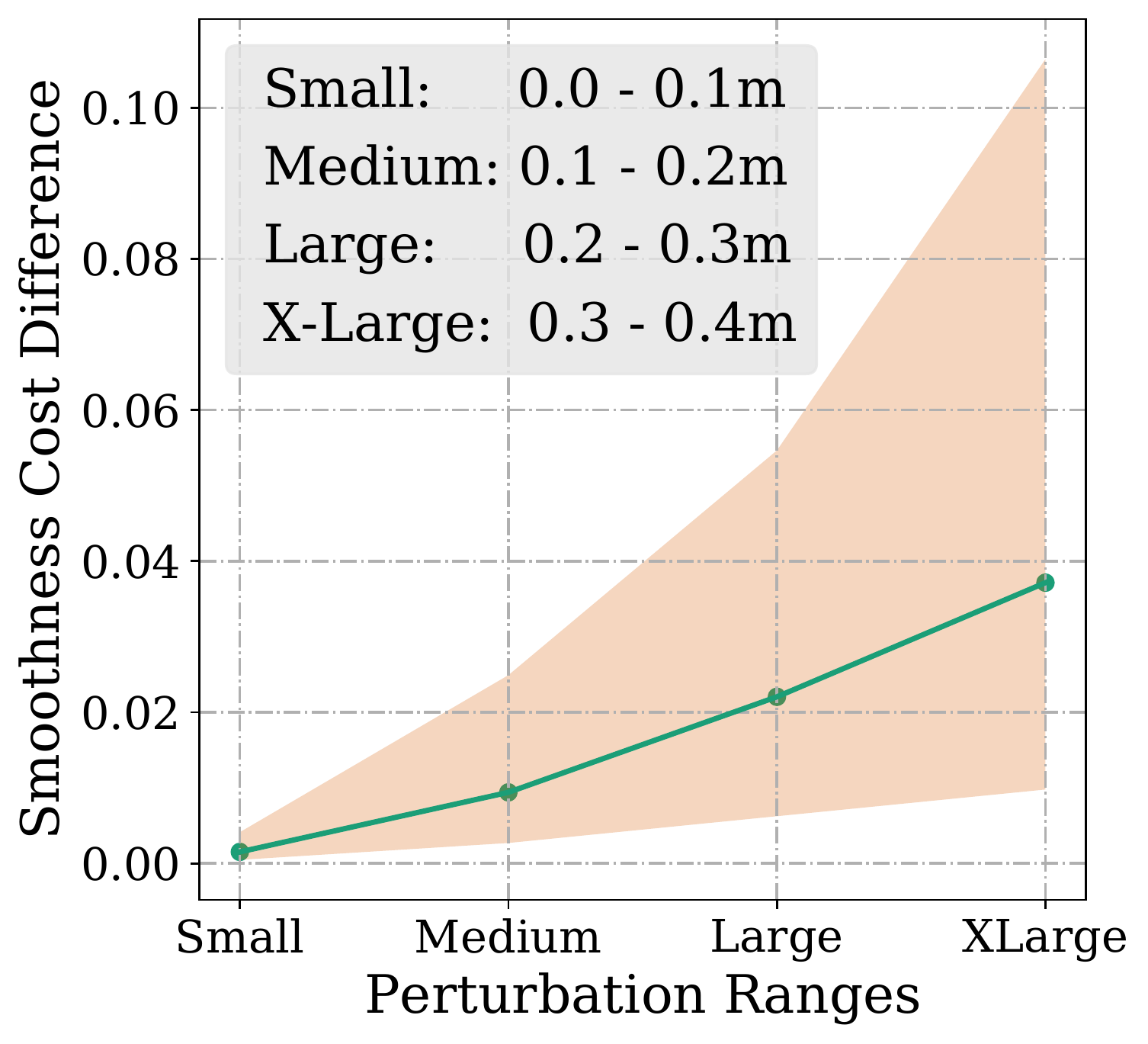}}
\subfigure{\label{fig:e4}\includegraphics[width=0.47\columnwidth]{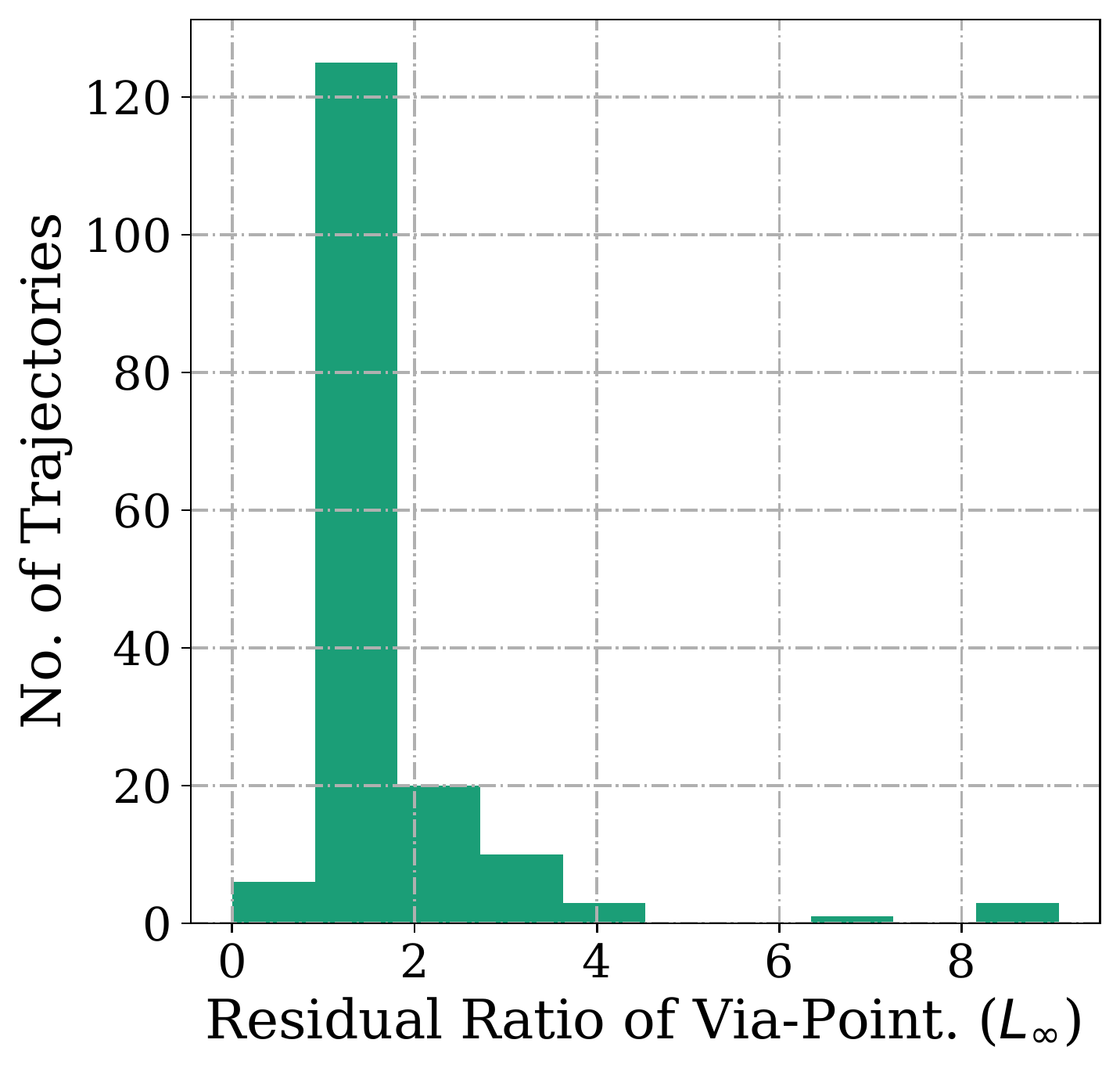}}
\subfigure{\label{fig:f4}\includegraphics[width=0.51\columnwidth]{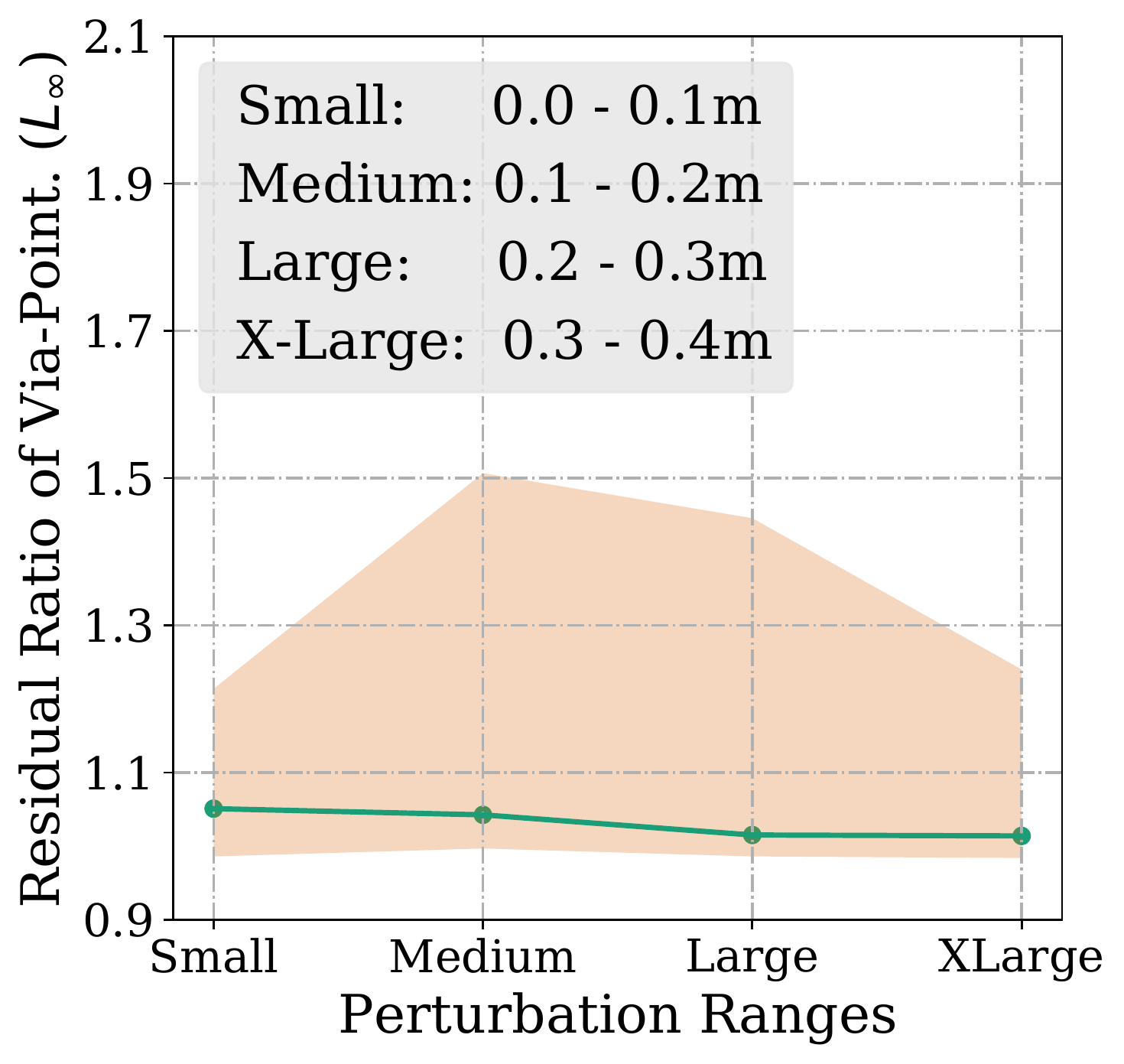}}
\caption{Performance of Algorithm \ref{algo_1} for different perturbation ranges on the benchmark of Fig. \ref{via_point} that involves perturbing the via-point of the end-effector trajectory (recall cost function (\ref{pos_interpol_cost})). The left and right columns show similar benchmarking as those of Fig. \ref{final_joint_pert}.}
\label{via_point_pert}
\end{figure}

% The latter is required for benchmarking with Algorithm \ref{algo_1}.

Each joint trajectory is parameterized by a 50-dimensional vector of way-points. Thus, the underlying task constrained trajectory optimization involves a total of $350$ variables. We use Scipy-SLSQP \cite{scipy} to obtain the prior trajectory and also to re-solve the trajectory optimization for the perturbed parameters. We did our implementation in Python using Jax-Numpy \cite{jax} to compute the necessary Jacobian and Hessian matrices. We also used the just-in-time compilation ability of JAX to create an on-the-fly complied version of our codes. The line-search in Algorithm \ref{algo_1} (line 2) was done through a parallelized search over a set of discretized $\eta$ values. The entire implementation was done on a 32 GB RAM i7-8750 desktop with RTX 2080 GPU (8GB). To foster further research in this field and ensure reproducibility, we open-source our implementation for review at \url{https://rebrand.ly/argmin-planner}

\begin{figure}[!h]
\centering     %%% not \center
\subfigure{\label{fig:a1}\includegraphics[width=0.47\columnwidth]{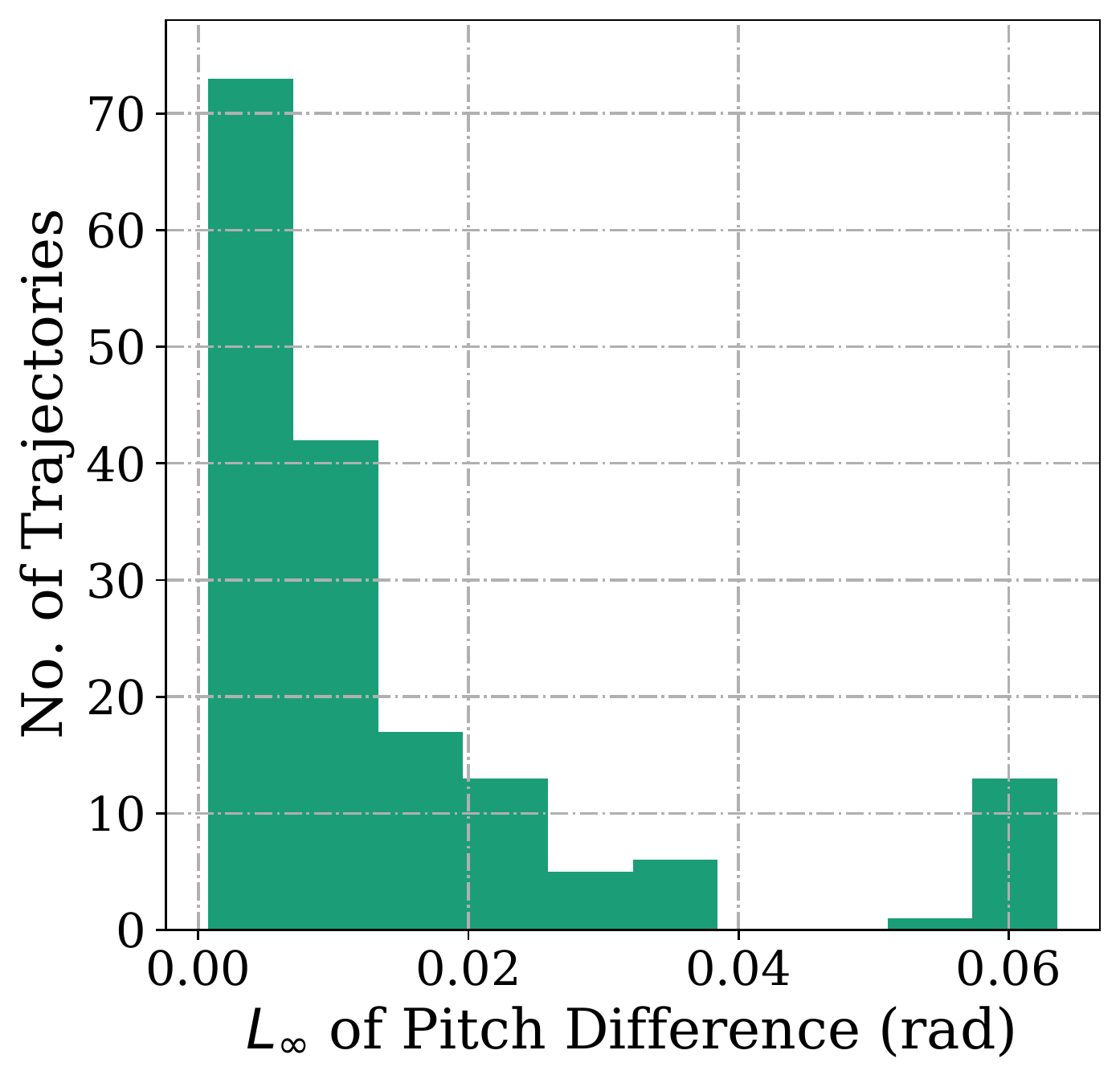}}
\subfigure{\label{fig:b1}\includegraphics[width=0.51\columnwidth]{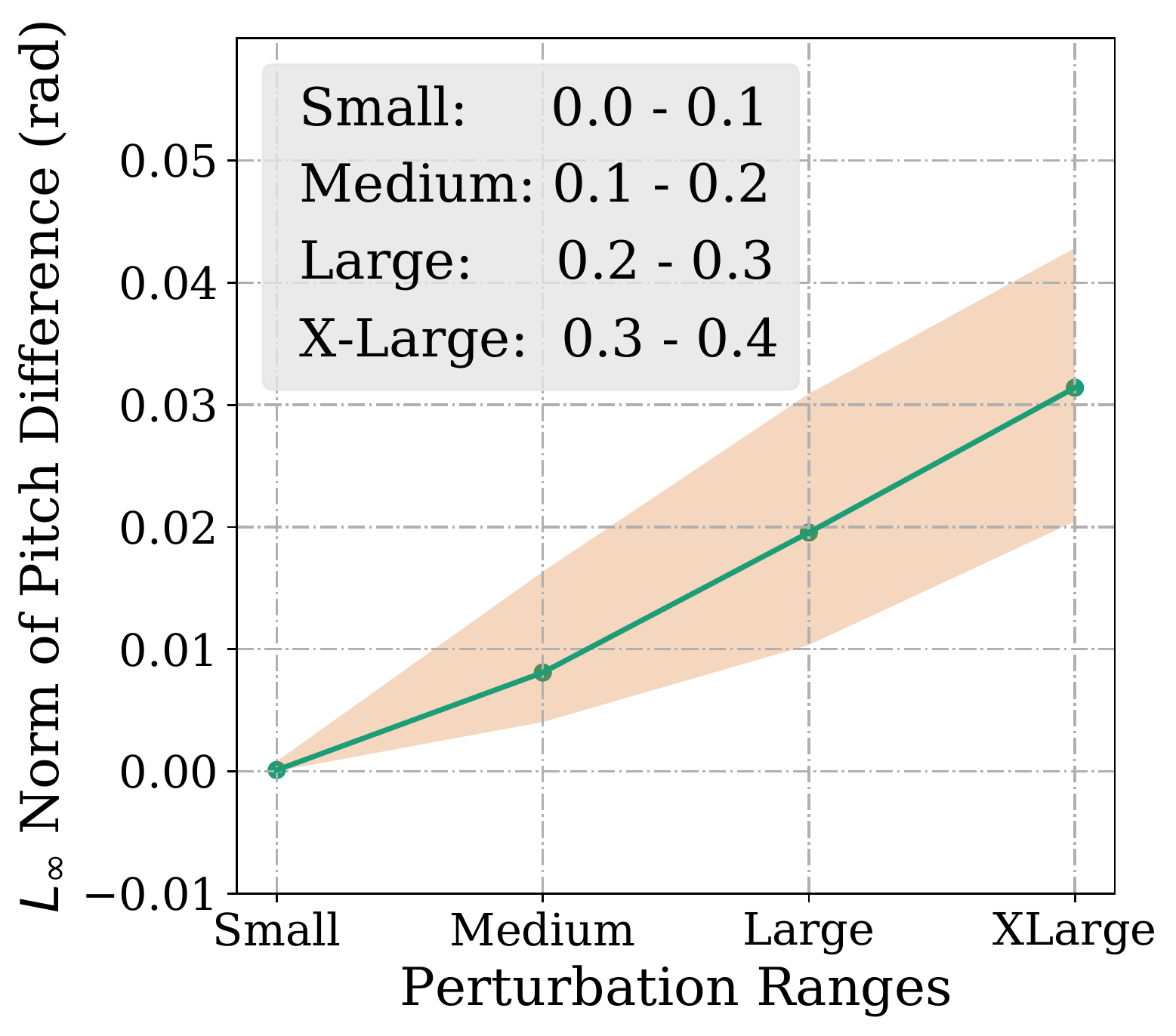}}
\subfigure{\label{fig:a1}\includegraphics[width=0.47\columnwidth]{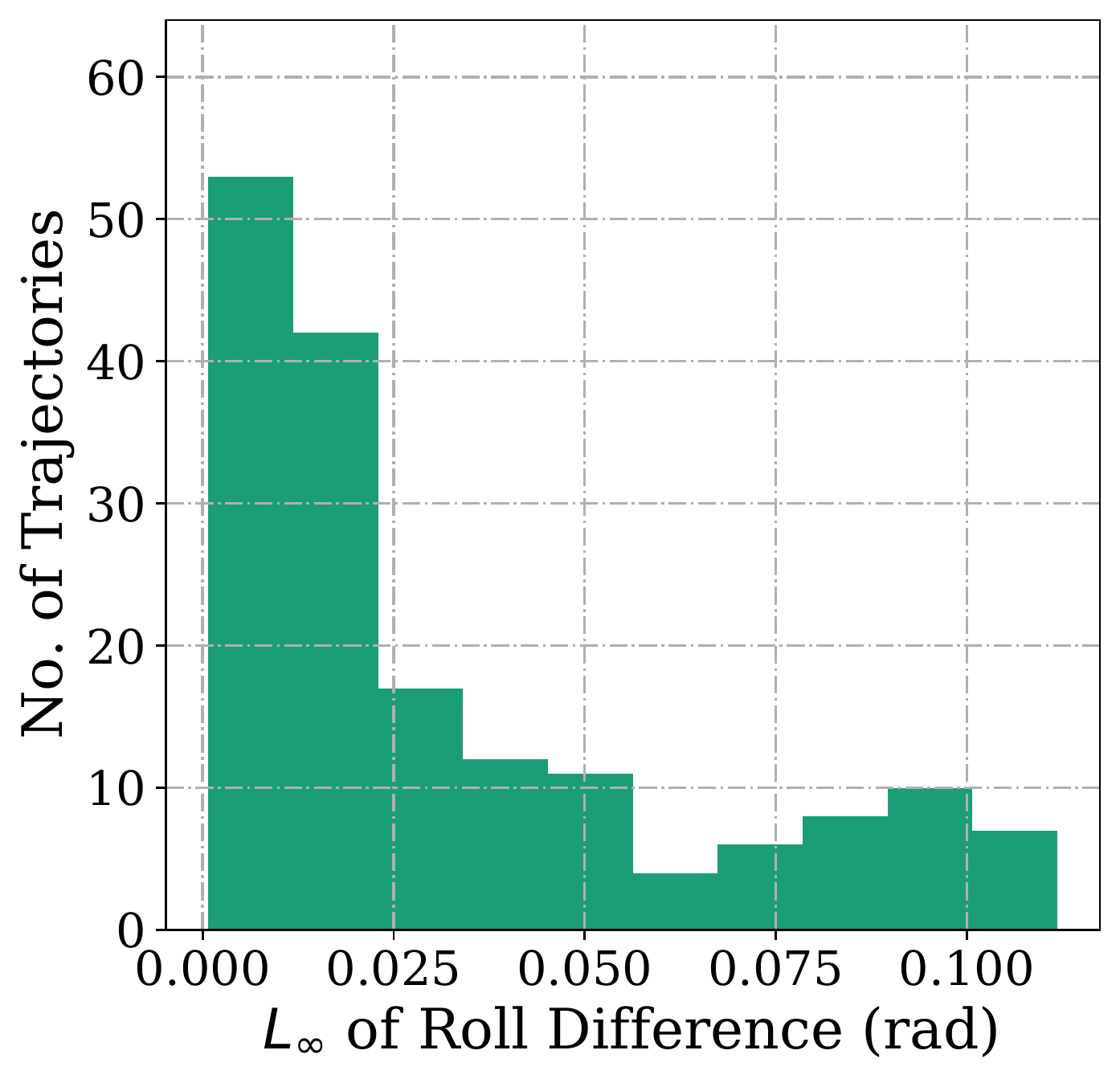}}
\subfigure{\label{fig:b1}\includegraphics[width=0.51\columnwidth]{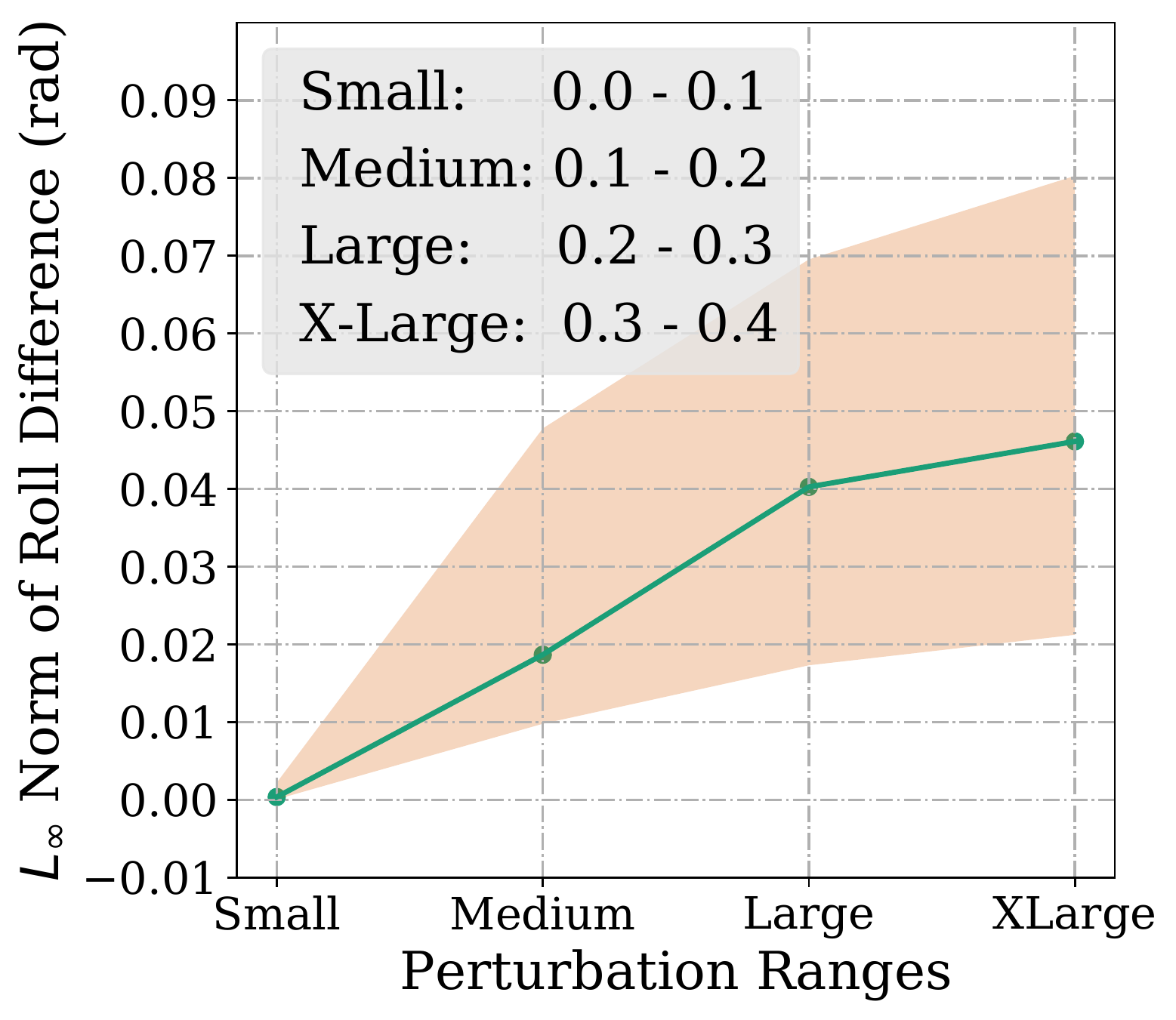}}
\subfigure{\label{fig:a1}\includegraphics[width=0.47\columnwidth]{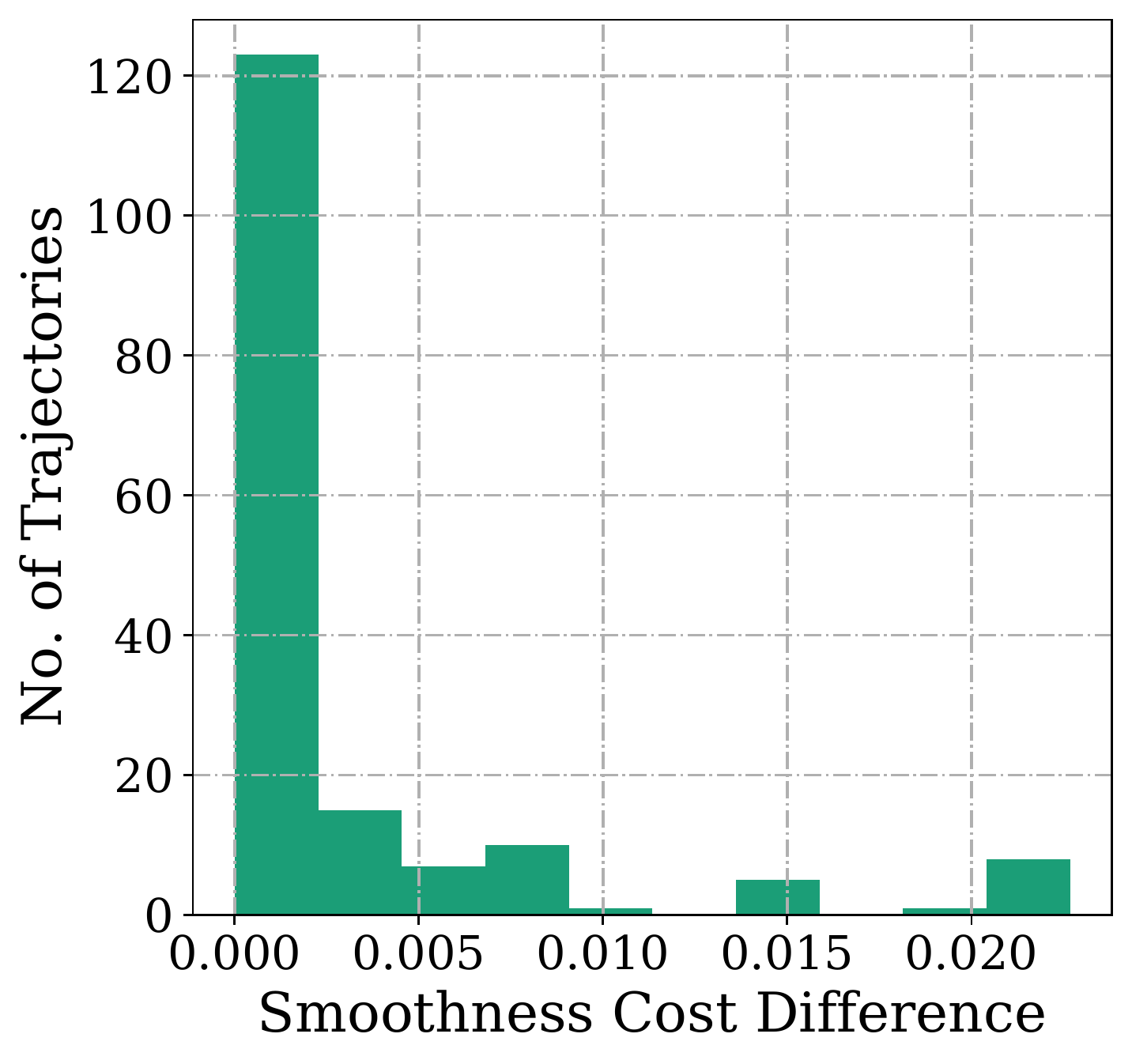}}
\subfigure{\label{fig:b1}\includegraphics[width=0.51\columnwidth]{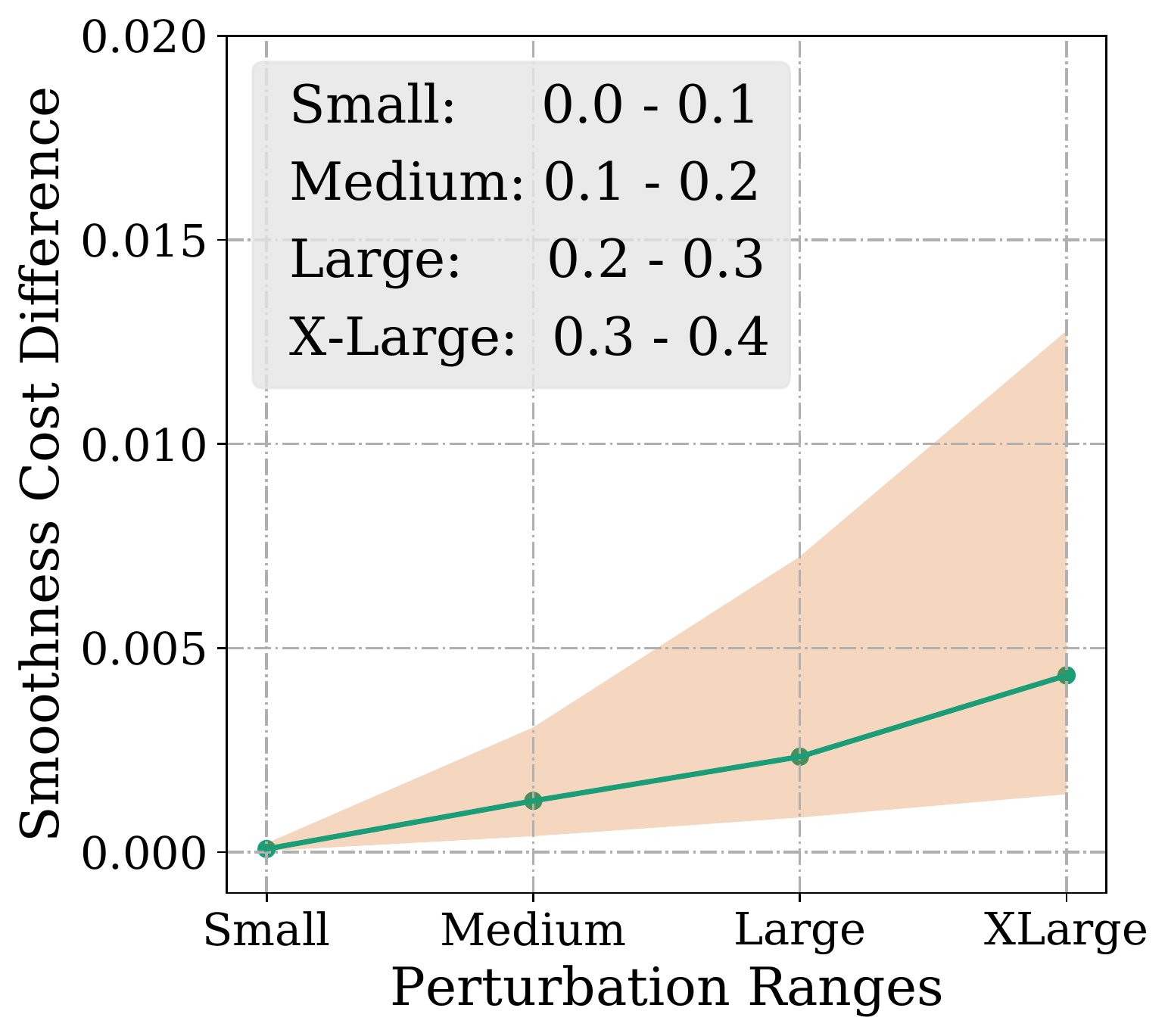}}
\subfigure{\label{fig:a1}\includegraphics[width=0.47\columnwidth]{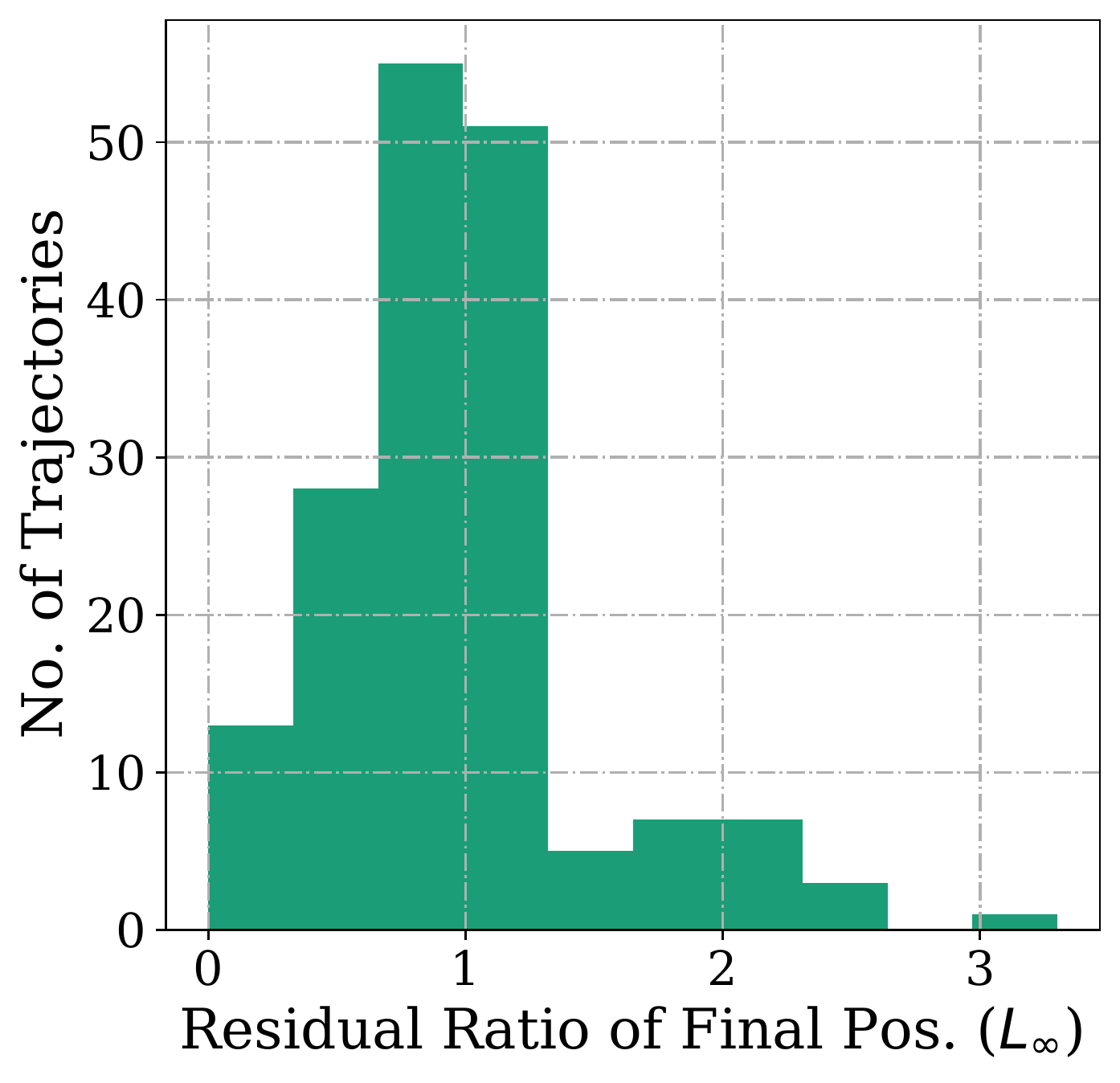}}
\subfigure{\label{fig:b1}\includegraphics[width=0.51\columnwidth]{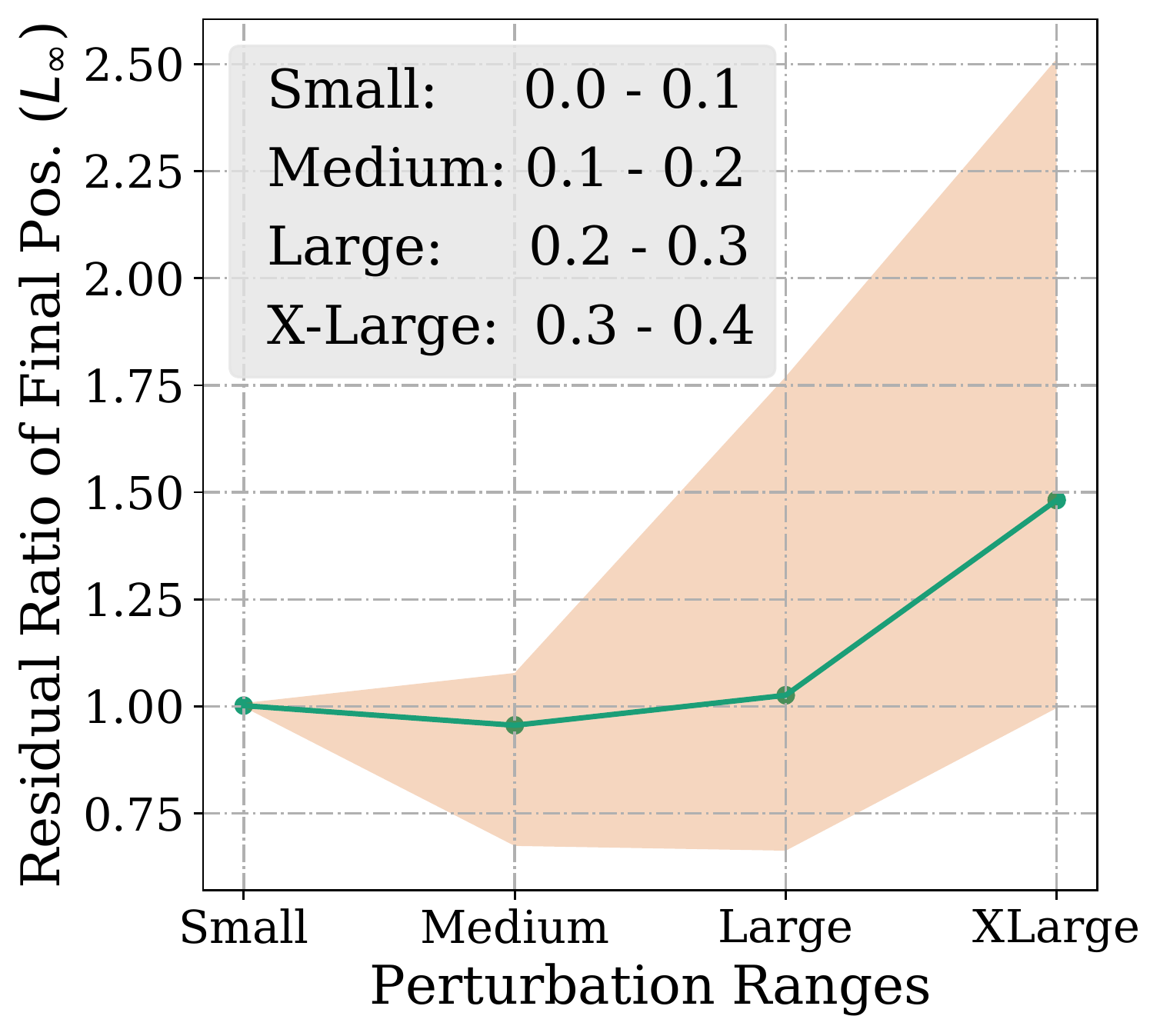}}
\caption{Performance of Algorithm \ref{algo_1} for different perturbation ranges on the benchmark of Fig. \ref{final_pos_pert} that involves perturbing the final end-effector position. (recall cost function (\ref{pos_interpol_cost})). The left and right columns show similar benchmarking as those of Fig. \ref{final_joint_pert}.}
\label{final_pos_pert}
\end{figure}

\subsection{Quantitative Results}
% We report the performance of our approach based on task, velocity smoothness of the computed trajectories, and the residual ratio of parameters for a given problem. We describe these metrics in detail below and report the results in Fig. \ref{final_joint_pert}-\ref{final_pos_pert} respectively. Our evaluation in Fig. \ref{final_joint_pert}-\ref{final_pos_pert} is reported on $4$ datasets with varying level of perturbation and with an average of $180$ trajectories per dataset. The histogram plots in Fig. \ref{final_joint_pert}-\ref{final_pos_pert} are reported for the \textbf{medium} perturbation dataset and all the line plots in Fig. \ref{final_joint_pert}-\ref{final_pos_pert} represent the first quartile, median and the third quartile of the concerned metric respectively.

\noindent \textbf{Orientation Metric:} For this analysis, we compared the pitch and roll angles at each time instant along trajectories obtained with Algorithm \ref{algo_1} and the resolving approach. Specifically, we computed the maximum of the absolute difference (or $L_{\infty}$ norm) of the two orientation trajectories. The yaw orientation in all these benchmarks was a free variable and is thus not included in the analysis. The results are summarized in Fig. \ref{final_joint_pert}-\ref{final_pos_pert}. The histogram plot in these figures are generated for the medium perturbation ranges (note the figure legends). For the Fig. \ref{orient_interpol} benchmark related to cost function (\ref{orient_interpol_cost}), Fig. \ref{final_joint_pert} shows that all the trajectories obtained by Algorithm \ref{algo_1} have  $L_{\infty}$ norm of the orientation difference less than 0.1 rad. For the benchmark of Fig. \ref{via_point}, which we recall involves perturbing the via-point of the end-effector trajectory, the histograms of Fig. \ref{via_point_pert} show similar trends. All the trajectories computed by Algorithm \ref{algo_1} managed a similar orientation difference. For the benchmark of Fig. \ref{final_pos} pertaining to perturbation of the final position $69.41\%$ of the trajectories obtained by Algorithm \ref{algo_1} managed to maintain a orientation difference of 0.1 rad with the resolving approach. 

% For the benchmark of Fig. \ref{s_curve} that involves perturbing the entire end-effector trajectory, the performance degrades a little with $86.2\%$ of the Algorithm \ref{algo_1} computed trajectories achieving $L_{\infty}$ of 0.1 rad in the orientation difference.

\noindent \textbf{Task residuals ratio metric:} For this analysis, we compare the task residual between trajectories obtained from  Algorithm \ref{algo_1} and the resolving approach. For example, for the benchmark of Fig. \ref{orient_interpol}, we want that the manipulator final configuration should be close to the specified value (recall cost (\ref{orient_interpol_cost})) while maintaining desired orientation at each time instant. Thus, we compute the $L_{\infty}$ residual of $\textbf{q}_t-\textbf{q}_m$ for Algorithm \ref{algo_1} and compare it with that obtained from the resolving approach. Now, just like before and to be consistent with the other benchmarks, we convert the residual of joint angles to position values through forward kinematics. Similar analysis follow for the other benchmarks as well. For the ease of exposition, we divide the task residual of Algorithm \ref{algo_1} by that obtained with the resolving approach. A ratio greater than 1 implies that the former led to a higher task residual than the latter and vice-versa. Similarly, a ratio closer to 1 implies that both the approaches performed equally well.

The results are again summarized in Fig. \ref{final_joint_pert}-\ref{final_pos_pert}. From Fig. \ref{final_joint_pert}, we notice that $97.05\%$ of trajectories have residual ratio less than $1.2$. For the experiment involving via-point perturbation in Fig. \ref{via_point_pert}, the performance drops to $62.50\%$ for the same value of residual ratio. Meanwhile,  as shown in Fig. \ref{final_pos_pert}, around $82.94\%$ of the trajectories have residual ratio less than $1.2$ in the case of final position perturbation benchmark of Fig.\ref{final_pos}.

% computed the ratio of residuals ($L_{\infty}$ norm of parameters achieved by individual method and desired parameters). The value of this ratio indicates how close our solution is to the solver's solution with respect to parameter cost, i.e, ratio closer to 1 indicates our solution is equivalent to solver solution and $>1$ indicates our solution performing worse. These results are again summarized in Fig. \ref{final_joint_pert}-\ref{final_pos_pert}. From Fig. \ref{final_joint_pert}, we notice that $97.05\%$ of trajectories have residual ratio less than $1.2$. For experiment involving viapoint perturbation in Fig. \ref{via_point_pert}, the performance drops to $62.50\%$ for the same value of residual ratio.Meanwhile, around $82.94\%$ of the trajectories have residual ratio less than $1.2$ in the case of final point perturbation experiment as shown in Fig. \ref{final_pos_pert}. 

\noindent \textbf{Velocity Smoothness Metric:} For this analysis, we computed the difference of the velocity smoothness cost  ($L_2$ norm of first-order finite difference) between the trajectories obtained with Algorithm \ref{algo_1} and the resolving approach. The results are again summarized in Fig. \ref{final_joint_pert}-\ref{final_pos_pert}. For all the benchmarks, in around $65\%$ of the examples, the difference was less than 0.05. This is $35\%$ of the average smoothness cost observed across all the trajectories from both the approaches. 

\noindent \textbf{Scaling with Perturbation Magnitude:} The line plots in Fig. \ref{final_joint_pert}-\ref{final_pos_pert} represent the first quartile, median and the third quartile of the three metrics discussed above for different perturbation ranges. 

For the benchmark of Fig. \ref{orient_interpol}, trajectories from Algorithm \ref{algo_1} maintains an orientation difference of less than 0.1 rad with the trajectories of the resolving approach for perturbations as large as 40 cm. The difference of smoothness cost for the same range is also small with median value being in the order of $10^{-3}$. The median task residuals achieved by  Algorithm \ref{algo_1} is only $2\%$ higher than that obtained by the resolving approach. For the benchmark of Fig. \ref{via_point}, the performance remains same on the orientation metric but the median difference in smoothness cost and task residual ration increases to 0.04  and $9\%$for the largest perturbation range. The benchmark of Fig. \ref{final_pos_pert} follow similar trend in orientation and smoothness metric but performs significantly worse in task residuals. For the largest perturbation range, Algorithm \ref{algo_1} leads to $50 \%$ higher median task residuals. However, importantly, for perturbation up to 30 cm, the task residual ratio is close to 1 suggesting that Algorithm \ref{algo_1} performed as good as the resolving approach for these perturbations.

% \begin{figure}[]
% \centering     %%% not \center
% \subfigure{\label{fig:a3}\includegraphics[width=0.47\columnwidth]{figs_icra/beizer/pitch_cost.pdf}}
% \subfigure{\label{fig:b3}\includegraphics[width=0.51\columnwidth]{figs_icra/beizer/pitch_all.pdf}}
% \subfigure{\label{fig:c3}\includegraphics[width=0.47\columnwidth]{figs_icra/beizer/roll_cost.pdf}}
% \subfigure{\label{fig:d3}\includegraphics[width=0.51\columnwidth]{figs_icra/beizer/roll_all.pdf}}
% \subfigure{\label{fig:e3}\includegraphics[width=0.47\columnwidth]{figs_icra/beizer/param_cost.pdf}}
% \subfigure{\label{fig:f3}\includegraphics[width=0.51\columnwidth]{figs_icra/beizer/param_all.pdf}}
% \caption{Performance of Algorithm \ref{algo_1} for different perturbation ranges on the benchmark of Fig. \ref{s_curve} that involves perturbing the complete end-effector trajectory (recall cost function (\ref{pos_interpol_cost})).  }
% \label{s_curve_pert}
% \end{figure}

\noindent \textbf{Computation Time:} Table \ref{table:dataset_description} contrasts the average timing of our Algorithm \ref{algo_1} with the approach of resolving the trajectory optimization with warm-start initialization. As can be seen, our Argmin differentiation based approach provides a worst-case speed up of 160x on the benchmark of Fig. \ref{final_pos}. For the rest of the benchmarks, this number varies between 500 to 1000. We believe that this massive gain in computation time offsets whatever little performance degradation in terms of orientation, smoothness, and task residual metric that Algorithm \ref{algo_1} incurs compared to the resolving approach. Note that the high computation time of the resolving approach is expected, given that we are solving a difficult non-convex function with 350 decision variables. Even highly optimized planners like \cite{dimitry_manifold_planning} show similar timings on closely related benchmarks \cite{const_x}.  

% \shashank{Nevertheless}, we open-source our implementation for review in \url{https://rebrand.ly/argmin-planner} 
% \textcolor{red}{github rep link}. 

\begin{table}[!hbt]
	\centering
    \begin{adjustbox}{max width=\linewidth}
	\begin{tabular}{|l|c|c|c|}
	\hline
	 & \multicolumn{2}{c|}{SciPy-SLSQP} & \multicolumn{1}{c|}{Ours Algorithm \ref{algo_1} }\\
    \cline{2-4}
     \multicolumn{1}{|c|}{Benchmarks} &\makecell{Wall time \\\textbf{(s)}} & \makecell{Wall time w/o \\jacobian and \\ function evaluation \\ overhead \textbf{(s)}} & \makecell{Wall time\\\textbf{(s)}} \\
    \hline
    Final Configuration Perturbation (Fig. \ref{orient_interpol}) & $43.91$ & $41.09$ & $\mathbf{0.039}$\\
    Via Point Perturbation (Fig. \ref{via_point})     & $53.05$ & $34.74$ & $\mathbf{0.09}$\\
    Final Position Perturbation (Fig. \ref{final_pos})   & $35.91$ & $29.09$ & $\mathbf{0.18}$\\
    % Complete trajectory perturbation (Fig. \ref{s_curve})   & $52.02$  & $38.57$ & $\mathbf{0.049}$\\
    \hline
	\end{tabular}
    \end{adjustbox}
    % \\[5pt]    
    \caption{Computation times comparison between Algorithm \ref{algo_1} and resolving trajectory optimization approach on three benchmarks.}
    \label{table:dataset_description}
\end{table}

\section{Conclusions and Future Work}
We presented a fast, near real-time algorithm for adapting joint trajectories to task perturbation as high as 40 cm in the end-effector position, almost half the radius of the Franka Panda arm's horizontal workspace used in our experiments. By consistently producing trajectories similar to those obtained by resolving the trajectory optimization problem but in a small fraction of a time, our Algorithm \ref{algo_1} opens up exciting possibilities for reactive motion control of manipulators in applications like human-robot handover.

In future works, we will extend our formulation to problem with dynamic constraints such as torque bounds. We conjecture that by coupling the way-point parametrization with a multiple-shooting like approach, we can retain the constraints as simple box-bounds on the decision variables and consequently retain the computational structure of the Algorithm \ref{algo_1}. We are also currently evaluating our Algorithm's performance on applications such as autonomous driving.

\newpage

% To this end, we define the 

% We begin by formalizing the task constrained trajectory optimization under the assumption 

% Adopting way-point parametrization for the joint trajectories, the optimization problem can be formulated in the following manner.

\bibliographystyle{IEEEtran}  
\bibliography{icra21_ref} 

\end{document}